\documentclass[journal]{IEEEtran}
\usepackage{amsmath,amsfonts}
\usepackage{algorithmic}
\usepackage{algorithm}
\usepackage{array}
\usepackage[caption=false,font=scriptsize,labelfont=sf,textfont=sf]{subfig}
\usepackage{textcomp}
\usepackage{url}
\usepackage{verbatim}
\usepackage{graphicx}
\usepackage{cite}

\usepackage{microtype}
\usepackage{booktabs}
\usepackage{xspace}
\usepackage{siunitx}
\usepackage{pstricks}
\usepackage[misc,geometry]{ifsym} 
\usepackage{hyperref}
\usepackage[capitalise]{cleveref}

\makeatletter
\DeclareRobustCommand\onedot{\futurelet\@let@token\@onedot}
\def\@onedot{\ifx\@let@token.\else.\null\fi\xspace}

\def\eg{\emph{e.g}\onedot} 
\def\ie{\emph{i.e}\onedot}

\def\etal{\emph{et al}\onedot}
\makeatother

\begin{document}

\title{Coarse-to-Fine Non-rigid Multi-modal Image Registration for Historical Panel Paintings based on Crack Structures}

\author{Aline Sindel \qquad Andreas Maier \qquad Vincent Christlein
\thanks{Received 31 December 2025}
\thanks{Thanks to Wibke Ottweiler, Oliver Mack, and Daniel Hess, Germanisches Nationalmuseum Nürnberg for providing the image data and for their art technological expertise.}
\thanks{The authors gratefully acknowledge the scientific support and HPC resources provided by the Erlangen National High Performance Computing Center (NHR@FAU) of the Friedrich-Alexander-Universität Erlangen-Nürnberg (FAU). The hardware is funded by the German Research Foundation (DFG).}
\thanks{Thanks to NVIDIA for their donation of one Titan Xp GPU.}
\thanks{(Corresponding author: Aline Sindel)}
\thanks{A. Sindel, A. Maier, and V. Christlein are with the Pattern Recognition Lab, Friedrich-Alexander-Universität Erlangen-Nürnberg, Erlangen, Germany.}%
\thanks{The code and trained models will be available after publishing.}}

\maketitle

\begin{abstract}
Art technological investigations of historical panel paintings rely on acquiring multi-modal image data, including visual light photography, infrared reflectography, ultraviolet fluorescence photography, x-radiography, and macro photography. For a comprehensive analysis, the multi-modal images require pixel-wise alignment, which is still often performed manually. Multi-modal image registration can reduce this laborious manual work, is substantially faster, and enables higher precision. Due to varying image resolutions, huge image sizes, non-rigid distortions, and modality-dependent image content, registration is challenging. Therefore, we propose a coarse-to-fine non-rigid multi-modal registration method efficiently relying on sparse keypoints and thin-plate-splines. Historical paintings exhibit a fine crack pattern, called craquelure, on the paint layer, which is captured by all image systems and is well-suited as a feature for registration. In our one-stage non-rigid registration approach, we employ a convolutional neural network for joint keypoint detection and description based on the craquelure and a graph neural network for descriptor matching in a patch-based manner, and filter matches based on homography reprojection errors in local areas. For coarse-to-fine registration, we introduce a novel multi-level keypoint refinement approach to register mixed-resolution images up to the highest resolution. We created a multi-modal dataset of panel paintings with a high number of keypoint annotations, and a large test set comprising five multi-modal domains and varying image resolutions. The ablation study demonstrates the effectiveness of all modules of our refinement method. Our proposed approaches achieve the best registration results compared to competing keypoint and dense matching methods and refinement methods.
\end{abstract}


\section{Introduction}
Art technological investigations of historical panel paintings are conducted to investigate, for instance, the material, alterations to the paintings, and state of preservation. Usually, multiple imaging systems, such as visual light photography (VIS), infrared reflectography (IRR), ultraviolet fluorescence photography (UV), x-radiography (XR), and macro photography (MACRO), are utilized. VIS provides an overview image of the painting by typically depicting the content in one piece. 
IRR is applied to reveal underdrawings beneath the layers of paint, UV to visualize overpaintings, restorations, and aged varnish, and XR is applied to provide information about the wooden panel and to highlight white lead. MACRO is used to analyze specific details of a painting in high resolution.
To connect and analyze the information from different imaging systems, the multi-modal images are overlaid. However, this requires alignment of the images, often conducted manually by the art technologists, since each modality is acquired using a different image resolution and potentially under slightly varying viewpoints, as they are not applied simultaneously.
Automatic multi-modal image registration methods are desirable, as they can reduce the tedious manual work, are much faster, and can achieve higher precision.

Multi-modal image registration approaches for panel paintings must tackle several challenges. On the one hand, the technical images of large panel paintings can be very large and require specialized algorithms for processing. 
On the other hand, mixed image resolution complicates feature matching. The VIS (and UV) overview images are usually lower-resolution because the complete painting must fit within the imaging view. In comparison, the other modalities are acquired in pieces at higher resolution. For IRR, the image tiles have a standard resolution of $4096 \times 4096$ pixels, XR are mostly scanned as longitudinal stripes with a length up to $44000$ pixels.
In addition, the overview images frequently show distortions in the border areas that cannot be corrected with a global perspective transform alone, and the IRR image tiles show vertical distortions.
Due to overpaintings and underdrawings, visual features can differ across the multi-modal images. Yet, the old paint of the historical panel paintings shows a fine network of crack structures (``craquelure''), which is depicted by all the modalities.

In this paper, we propose a \textit{coarse-to-fine non-rigid multi-modal registration method} for the task of panel-painting alignment. To address the challenge of \textit{processing very large images}, we developed an efficient method using sparse data, \ie, keypoints, for non-rigid registration with a thin-plate-spline (TPS). Further, we present a novel \textit{coarse-to-fine} registration scheme, which is \textit{modular} and thus can be applied to both multi-modal image pairs with similar and mixed image resolutions, by either just executing the first registration stage or the full pipeline, including \textit{step-by-step keypoint refinement}. The final transformation should be elastic, but still as rigid as possible locally. Therefore, we can use a global transformation, \ie, homography, to filter the point correspondences within local areas. 
To extract and match reliable keypoints and descriptors from all multi-modal images, we employ our previous work CraquelureNet~\cite{SindelA2021}, a \textit{keypoint and description network} focusing on the crack structures in the paint, and combine it with the \textit{deep feature matcher} LightGlue~\cite{LindenbergerP2023}. Further, we introduce novel \textit{refinement modules} to revise keypoint locations, and we evaluate our proposed methods on a \textit{large test set of panel paintings} spanning five multi-modal domains and varying resolutions.

\section{Related Work}

\subsection{Feature-based Registration}
Feature-based registration usually consists of the extraction of keypoints and descriptors, descriptor matching, and transformation estimation using the matched correspondences.

\subsubsection{Keypoint Detection and Description}
Early works purely use handcrafted features such as SIFT~\cite{LoweDG2004} or HOG~\cite{DalalN2005}.
With advances in deep learning, more and more parts of the traditional pipeline are replaced by using neural networks. 

A group of methods concentrated on developing strategies for descriptor learning using convolutional neural networks (CNNs). Image patches, typically $32 \times 32$ pixels, are fed to the CNN, which encodes each patch as a feature vector, the descriptor. Approaches using contrastive loss split the training data into positive and negative patch pairs, and minimize the distances of positive pairs and maximize the distances of negative pairs.
DeepDesc~\cite{SimoSerraE2015} trains CNN-based descriptors using contrastive loss and hard-negative and positive mining.
Triplet loss for descriptor learning~\cite{AltwaijryH2016,BalntasV2016,MishchukA2017,MishkinD2018} computes for an anchor the distance to the positive counterpart and to the negative counterpart, with the training objective to pull the positive counterpart closer to the anchor (minimize distance) and at the same time push the negative counterpart further apart (maximize distance). The triplets can be selected using classical online hard negative mining, \ie, closest negatives within the batch~\cite{AltwaijryH2016}, or using more complex strategies~\cite{BalntasV2016,MishchukA2017,MishkinD2018}, such as swaping the role of anchor and positive counterpart when the positive is closer to the negative than the anchor~\cite{BalntasV2016} or selecting the hardest non-matching descriptors for both anchor and positive counterpart and choosing then the one with the smaller distance~\cite{MishchukA2017}, or combining triplet loss and contrastive positive loss~\cite{MishkinD2018}. Inspired by these approaches, we extended in~\cite{SindelA2021} the triplet loss to a bidirectional quadruplet loss by using in-batch hardest non-matching descriptors for both anchor and positive counterpart.

Another group of methods applies self-supervised schemes to jointly train keypoint detection and description using ground-truth transformations.
SuperPoint~\cite{DeToneD2018} uses a shared VGG-based encoder and specific decoder heads for detection and description. The detector head computes softmax for the point scores in $8\times8$ cells. The descriptors are computed semi-densely and are bicubically interpolated at the keypoint positions.
Both heads are jointly trained using pseudo ground-truth interest points and the self-supervised technique homographic adaptation, \ie, an image is augmented into an image pair by applying a random homography, which is used as guidance for network training.
To generate the pseudo ground truth, a CNN is trained on synthetic shapes with labels at the interest points, \eg, junctions and endpoints of shapes or lines.
D2-Net~\cite{DusmanuM2019}, in particular, developed for difficult imaging conditions, \eg, day-night scenes, simultaneously learns the descriptor and detector with a single CNN that shares all parameters. The descriptors are feature vectors along the channel dimension of the multi-dimensional score map, and keypoints are extracted from all levels of the score map, but must be local maxima at the specific level and along the channel dimension.
R2D2~\cite{RevaudJ2019} focuses on keypoint repeatability and reliability and thus predicts two confidence maps, one for each task, and a dense descriptor map.
Disk~\cite{TyszkiewiczM2020} applies reinforcement learning to jointly learn a keypoint heatmap and dense descriptors using policy gradient.
Aliked~\cite{ZhaoX2023} uses a shared network for feature encoding and aggregation of different feature map levels and specific heads for detection and description. The keypoint extraction, adopted from their previous work~\cite{ZhaoX2023Alike}, is partially differentiable because it refines keypoint positions in patches around each keypoint using softargmax. The authors propose deformable descriptors that apply deformable convolution and learn, for each keypoint, supporting features at deformable locations in a sparse manner, rather than extracting descriptors from a dense feature map.

\subsubsection{Learning-based Feature Matching}
Classical descriptor matching, such as nearest neighbor (NN) or mutual NN matching (MNN), mostly combined with RANSAC~\cite{FischlerMA1981} for robust transformation estimation despite existing outliers, can be replaced or complemented with learning-based approaches.
Yi \etal~\cite{YiK2018CVPR} uses a set of putative correspondences obtained with NN descriptor matching as a starting point. They propose a multilayer perceptron with context normalization to predict, for each correspondence, whether it is an inlier or outlier match. For the optimization, the correspondence scores are used as weights for the essential matrix estimation.
The attentional graph neural networks, SuperGlue~\cite{SarlinPE2020} and its successor LightGlue~\cite{LindenbergerP2023}, model keypoint-based image matching as a partial assignment problem and predict a matching score matrix for all points using keypoints and descriptors.

Detector-free deep matchers directly estimate dense descriptors and matches without relying on keypoint positions, aiming to detect matches in low-texture regions as well.
LoFTR~\cite{SunJ2021} estimates dense matches at coarse level ($1/8$ resolution) and refines them at fine level ($1/2$ resolution) using ResNet-based features and a LoFTR module consisting of a Transformer with linear self- and cross-attention. The matching probability for coarse matching is estimated using double-softmax. Good matches are refined using a correlation-based LoFTR fine module. 
Several follow-ups have been proposed. 
MatchFormer~\cite{WangQ2022} replaces the CNN backbone with a Transformer using interleaving self- and cross-attention.
ASpanFormer~\cite{ChenH2022} introduces iterative global-local attention.
CasMTR~\cite{CaoC2023} adds cascade modules and non-maximum suppression (NMS).

Dense matchers, such as DKM~\cite{EdstedtJ2023} and Roma~\cite{EdstedtJ2024}, estimate a warping function and matchablitity score for each pixel in the two images.
Both methods first perform global matching using coarse features. 
Secondly, they iteratively refine the estimated coarse warp and confidence using a set of warp refiner modules with fine features, which upsample the warps and use local correlation of the warped feature maps.
Finally, they sample the correspondences for geometry estimation from the dense matches via balanced sampling. Roma~\cite{EdstedtJ2024} makes architectural changes to DKM~\cite{EdstedtJ2023}, such as using separate networks for the feature encoders or replacing the CNN-based match decoder with a Transformer-based one, and changes the loss functions. 

\subsection{Registration for Art Imaging}
Image registration methods for art imaging can be grouped into intensity-, control point-, and feature-based approaches.
In the area of intensity-based approaches mutual information was applied to iteratively register multispectral images of paintings~\cite{CappelliniV2005} or X-ray fluorescence and visual images of paintings~\cite{VillafaneME2023,AmiriMM2025}.

Control-point- and feature-based methods compute a geometric transformation based on correspondences between the images.
Control-point-based methods usually apply classical image processing techniques, \eg, in~\cite{MurashovD2011} local grayscale maxima extraction is combined with coherent point drift~\cite{MyronenkoA2010}.
The mosaicking approach by Conover \etal~\cite{ConoverDM2015} detects control-points in multi-modal images of paintings (VIS with IR and XR) using Wavelet transform, phase correlation, and spatial disparity filtering. Application-wise their approach is similar to ours, as they compute bilinear transformations for sub-images rather than a global transformation for the complete painting. However, they tessalate the individually transformed pieces to build the transformed image, while we use the regions only for point matching and filtering and compute a TPS using the displacement vectors for the complete image. 

Further, handcrafted features such as SIFT~\cite{LoweDG2004} and SURF have been applied to align multispectral images of artworks~\cite{ZacharopoulosA2018} or multi-modal images of wall paintings~\cite{AnzidH2023}.   

Our previous work, CraquelureNet~\cite{SindelA2021}, is the first multi-modal registration method that uses a CNN to extract keypoints and descriptors based on cracks, which are visible in the paint of historic paintings. Cracks serve as a constant feature throughout multiple imaging systems, while the depicted image content can vary due to underdrawings and overpaintings which are revealed by only some of the modalities. 

\subsection{Crack Detection}
Related work on crack detection in paintings, primarily focused on crack segmentation for virtual in-painting of cracks, crack-type classification, or to support restoration. These approaches use morphological operators~\cite{AbasFS2003,Spagnolo2010,DeborahH2015,RucobaC2022}, Bayesian conditional tensor factorization~\cite{CornelisB2013} or CNNs~\cite{SizyakinR2020}.

\subsection{Keypoint Refinement}
Recently, classical approaches to refine keypoints for image matching were investigated by Bellavia \etal~\cite{BellaviaF2024}, which are applied to patches around the keypoints. Template matching estimates an offset vector for the keypoint. 
Different cost functions, such as normalized cross-correlation (NCC), adaptive least-squares (ALS) correlation matching, and fast affine template matching (FAsT-Match), are evaluated. Additionally, sub-pixel refinements such as parabolic or Taylor-approximation peak interpolation are applied to the NCC response map.

Some detectors or dense matchers address keypoint refinement using patches at the keypoint or match position.
Patch2Pix~\cite{ZhouQ2021} regresses a refined match position with a CNN that takes as input both patches of the match.
Detectors like~\cite{OnoY2018,ZhaoX2023Alike,ZhaoX2023,SindelA2022MICCAI} use the softargmax operator for patches of the score map for sub-pixel refinement, though independently for each image.
The LoFTR fine matching module uses correlation of the feature vector and feature map patch to obtain the sub-pixel offset.
With this, LoFTR only refines keypoints in one of the two images.

KeyPt2SubPx~\cite{KimS2024} uses a CNN to refine arbitrary keypoint detectors for sub-pixel accuracy. The CNN transforms the stacked patches of the image and the score map around the keypoint into a feature map. The feature maps are enriched by applying the dot product with the mean of both descriptors. The offset vector is obtained by softargmax. During training, the CNN is optimized using ground-truth essential matrices. 

GMM-IKRS~\cite{SantellaniE2024} fits a Gaussian mixture model (GMM) to keypoints and their reprojections from multiple generated views of the input image, where the means of the GMM components are then the refined points.
However, due to its multiple image warping scheme, runtime would be drastically increased if applied to our large images.

\section{One stage non-rigid registration}
In this section, our multi-modal non-rigid registration pipeline is described, which can be applied alone or be the first stage of our coarse-to-fine registration pipeline.

\begin{figure}[!t]
\centering
\includegraphics[width=3.5in]{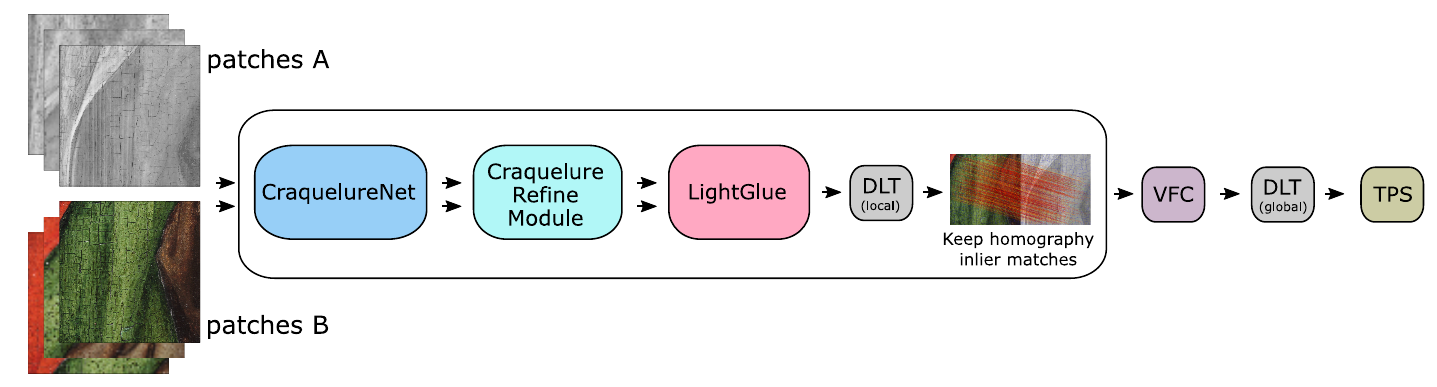} 
\caption{Our one-stage multi-modal non-rigid registration pipeline works on patches to detect good correspondences in local areas based on homography reprojection errors. Crack-based keypoints and descriptors are extracted using CraquelureNet and are refined to sub-pixel using our Craquelure Refine module. LightGlue is applied for matching and weighted DLT for homography estimation. All inlier point pairs are collected for all patches, are filtered using VFC, and a global homography is computed. The displacement vectors between the warped source and target points are used to compute a TPS.}
\label{flowchart-stage1}
\end{figure}
\cref{flowchart-stage1} shows the main components and steps.
To estimate a non-rigid transformation that preserves a rigid transform in local areas, and secondly, due to the very large image sizes, we process the images in a patch-based manner. For an image patch pair that potentially overlaps, we extract crack-based keypoints and descriptors using CraquelureNet. Keypoint positions are then refined using the proposed Craquelure Refine module. Then, LightGlue is used to predict potential good matches based on the craquelure keypoints and descriptors. The quality of the matches is checked based on some criteria, such as the local homography reprojection error or the number of inliers. The point correspondences and matching scores of the patches that meet the quality criteria are collected, and then post-filtering is applied to remove duplicates or very close matches, and finally, a global homography is computed using the weighted direct linear transform (DLT). Displacement vectors between the transformed source points using the global homography and the detected target points are used to compute a thin-plate-spline (TPS).

\subsection{CraquelureNet}

\begin{figure}[!t]
\centering
\includegraphics[width=3.5in]{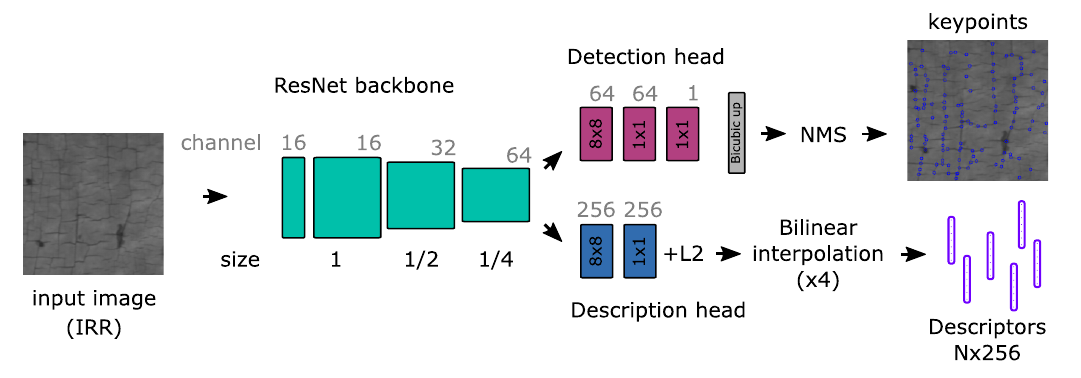}
\caption{CraquelureNet extracts keypoints and descriptors based on the crack structure in multi-modal images of historic paintings. It consists of a ResNet backbone and a detection and description head. The keypoint heatmap is bicubically upscaled and post-processed using non-maximum suppression (NMS). The descriptors are bilinearly interpolated at the keypoint locations in the L2-normalized descriptor map.}
\label{flowchart-CN}
\end{figure}

CraquelureNet~\cite{SindelA2021} is a keypoint detection and description network trained to focus on striking positions in the craquelure, e.g., a branching or sharp bend of a crack structure.
CraquelureNet consists of a ResNet backbone, a keypoint detection head, and a keypoint description head, for which we change the dimensions of the output channels, see~\cref{flowchart-CN}. The ResNet backbone with three layers of ResNet blocks, initially designed for the CIFAR-10 dataset, reduces the input image size by a factor of $4$. Both heads first apply $8 \times 8$ convolutional layers, followed by $1 \times 1$ convolutional layers. For keypoint extraction, the confidence heatmap is bicubically upscaled by a factor of $4$, and postprocessed with NMS. The descriptors are L2-normalized and linearly interpolated in the coarse descriptor grid using the upscaled keypoint positions. 

CraquelureNet is trained patch-based using small $32 \times 32 \times 3$ sized image patches and a multi-task loss for both heads:
\begin{equation}
\mathcal{L}_{\text{CN}} = \lambda_{\text{Det}} \mathcal{L}_{\text{BCE}} + \lambda_{\text{Desc}} \mathcal{L}_{\text{QuadB}},
\end{equation}
where $\lambda_{\text{Det}},\lambda_{\text{Desc}}$ are the weights for the detector and descriptor loss.
The keypoint detection head classifies the small patches into the two classes ``craquelure'' and ``background'' using binary cross-entropy loss $\mathcal{L}_{\text{BCE}}$.
The ``craquelure'' class is defined as a branching or sharp bend of a crack structure in the center of the patch. The ``background'' class contains cracks only in the patch's periphery, or none at all.
To learn cross-modal descriptors, the keypoint description head is trained using bidirectional quadruplet loss~\cite{SindelA2021} which applies an online hard-negative mining strategy:
\begin{align}
\begin{split}
\mathcal{L}_{\text{QuadB}}(a,p,n_a,n_p) &= \max [0, m + d(a,p) - d(a,n_a)] 
\\ &+ \max [0, m + d(p,a) - d(p,n_p)],
\end{split}
\end{align}
with the quadruplet $(a,p,n_a,n_p)$, margin $m$, and $d(x,y)$ being the Euclidean distance of the descriptors.
For a positive keypoint pair (anchor $a$ and positive $p$), in both directions the hardest non-matching descriptors ($n_a$ closest to $a$ and $n_p$ closest to $p$) are selected within the batch. The distances of the descriptors of the matching keypoint pairs are minimized, and simultaneously, the distances of the descriptors of the hardest non-matching keypoint pairs are maximized.

\subsection{Craquelure Refine module} \label{subsec_refinemodule}

\begin{figure}[!t]
\centering
\includegraphics[width=.35\textwidth]{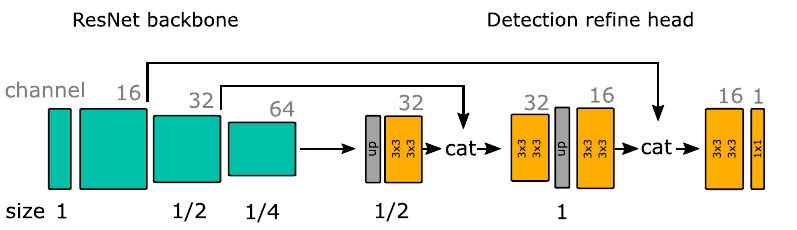}
\caption{The Craquelure Refine module extends the ResNet backbone of CraquelureNet with the Detection refine head to a U-Net for sub-pixel keypoint refinement.}
\label{flowchart-CNRefineModule}
\end{figure}

In the original setting of CraquelureNet, the output of the Craquelure detection head is upscaled by a factor of 4 for keypoint prediction to obtain a keypoint confidence heatmap at the full input image resolution. To refine these upscaled keypoint positions, we introduce a Craquelure refine module, which is plugged in directly after the keypoint extraction. 
 
The network architecture of the refine module is shown in~\cref{flowchart-CNRefineModule}. The pretrained ResNet backbone of CraquelureNet with frozen weights is extended by the Detection refine head with decoders to form a U-net-like encoder-decoder with a single output channel.
The Craquelure Refine module is trained in a supervised manner using true and noisy keypoint coordinates. Noisy keypoints are generated by shifting the true sub-pixel keypoint coordinates by a random offset within a range of $[-1.5,1.5]$ pixels. 
Then, $32 \times 32 \times 3$ patches are extracted at the rounded noisy keypoint centers and fed through the network. A $7 \times 7$ heatmap is interpolated at the sub-pixel noisy center of the $32 \times 32$ heatmap output. The learning task is to predict the true center in the heatmap using softargmax and is minimized using MSE loss.

\subsection{Graph neural networks for descriptor matching}
Descriptor matching is basically a partial assignment problem; a point in image A shall be assigned to only one point in image B. Since some points are unmatchable, it is usually formulated as a soft partial assignment problem.
The pioneering work SuperGlue~\cite{SarlinPE2020} solves the soft partial assignment problem with an attention-based graph neural network and the optimal transport formulation. 
In this work, we integrate LightGlue into our registration pipeline, which builds on SuperGlue, improving the accuracy and efficiency of the deep matcher.
LightGlue processes the keypoints and descriptors of both images through a set of layers with self- and cross-attention units and positional encoding. The final layer predicts the partial assignment matrix from pairwise similarity and matchability scores using double-softmax instead of the computationally more expensive optimal transport.

\subsection{Matching criteria and correspondences filtering} \label{subsec_filtering}
To determine whether the correspondences of a matched pair are sound and should be kept, we define the following criteria:
\begin{enumerate}
	\item Number of matches greater than 20
	\item Finding a valid homography (diagonal elements of homography not singular, small perspective parameters)
	\item Inliers w.r.t. homography greater than 10 (measured based on the reprojection error being smaller than a threshold)
\end{enumerate}

In the end, there can be duplicates or point pairs with very small distances to one another, since on the one hand patches are extracted overlappingly and on the other hand each patch of the first image is matched with multiple patches from the second image in a specified area, as the excerpt and scale of the image content are not necessarily exactly the same. We sort the points by match confidence and filter out nearby (duplicated) points, i.e., points within a given spatial distance threshold of the given point. 
Further, to filter out mismatches, we apply Vector field Consensus (VFC) filtering~\cite{MaJ2014}, which we reimplemented in Python using the authors' Matlab code.
VFC is a robust outlier removal method which interpolates a vector field for the correspondences and formulates the classification into outliers and inliers as a maximum a posteriori estimation using the EM algorithm.

\section{Coarse-to-fine non-rigid registration}
In this section, we propose a non-rigid registration method for large panel paintings with mixed resolutions, progressing step-by-step from a coarse scale to the finest scale of the images.
On the coarsest image resolution level, the one-stage non-rigid registration pipeline described in the previous section is applied to obtain a set of good point correspondences for the complete image pairs.

\subsection{High-resolution upscaling and keypoint refinement}
Given the low-resolution correspondences, we propose a two-step approach for keypoint upscaling and refinement. 
Ideally, keypoints should be centered at the central positions of the craquelure. By upscaling low-resolution keypoints to the finest resolution, small deviations of the keypoints from the center are magnified; hence, a step-by-step refinement is required to move the keypoints at each resolution level to the craquelure center.

For each image resolution level, first, the images and keypoints are upscaled to the next level. Images with an original resolution lower than the finest scale are upscaled, while high-resolution images are downscaled to the resolution of the current level.
The images are split into a grid of non-overlapping patches, such that each keypoint pair belongs to only one region, and the refinement is applied region-wise.
For each region, we first refine the modality with the higher resolution (modality 1), then shift the keypoints in the second modality to match those of the first.

\subsubsection{Craquelure score map-based refinement}
For the keypoint refinement in modality 1, we reuse and retrain the Craquelure Refine module (defined in \cref{subsec_refinemodule}), which predicts refined keypoint score maps for $32 \times 32$ patches centered at the upscaled keypoint positions. However, for the first refinement level, we want to give the keypoint greater movement flexibility. Thus, we introduce a search region around the upscaled keypoint of $18 \times 18$ pixels, within which we select the top $k=4$ scores from the score map and choose the spatially closest one to the original point. Around this updated point in a $7 \times 7$ region, we then estimate the softargmax to compute the sub-pixel offset. Conversely, at the second refinement level or if only one refinment level is applied, we skip the search region and compute the softargmax directly. 

\subsubsection{Craquelure feature volume correlation-based refinement}
\begin{figure}[!t]
\centering
\includegraphics[width=3.5in]{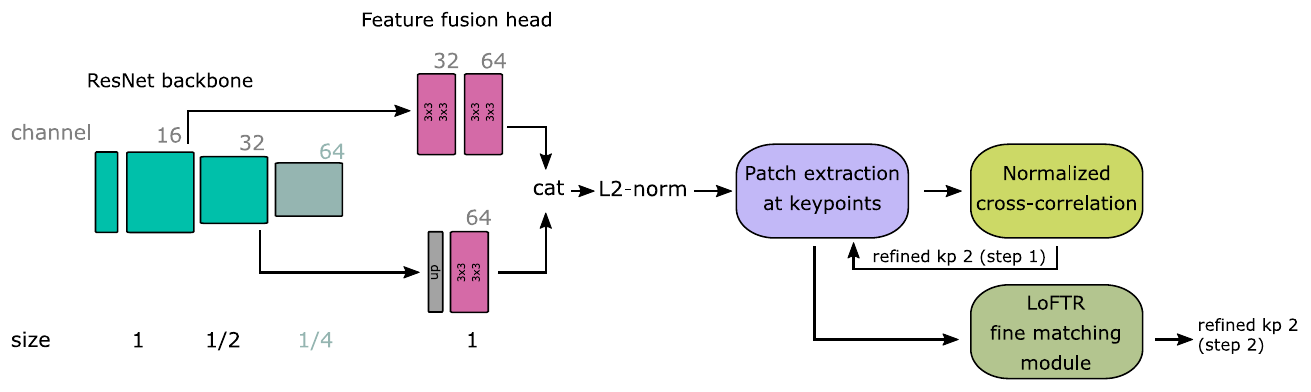}
\caption{Feature fusion head to extract craquelure features for normalized cross-correlation (NCC) refinement and for our adapted correlation-based LoFTR fine matching module.}
\label{flowchart-CNFeatRefineModule}
\end{figure}

We aim to shift the upscaled keypoints in modality 2 to the best matching position regarding the refined keypoints in modality 1.
We implement this by performing correlation of craquelure-specific feature volumes, first using normalized cross-correlation (NCC) for initial alignment, then using an adapted LoFTR fine matching correlation module for subsequent sub-pixel refinement. For this, we add a feature fusion head that takes the intermediate feature maps from the Craquelure backbone at the first and second ResNet blocks, as shown in~\cref{flowchart-CNFeatRefineModule}. The high-resolution feature maps pass through two convolutional blocks with two $3 \times 3$ convolutions each, Batch normalization, and Leaky ReLU to increase the feature dimension to $64$. The low-resolution feature maps are first bilinearly upsampled and then passed through a convolutional block similar to the former. The $64$-channel outputs from both paths are concatenated and L2-normalized. 
Since we defined a search region for modality 1, consequently, we also integrate one for modality 2. We perform NCC on a $24 \times 24$ region of the craquelure feature volumes, centered at refined keypoints in modality 1 and at upscaled keypoints in modality 2, at the sub-pixel level.
For each keypoint, the NCC produces a score map, where the highest score corresponds to the coordinate that best matches the keypoint in modality 1. Therefore, the refined keypoint is determined in the NCC score map by argmax and subsequent softargmax within a $7 \times 7$ region around the maximum.
Next, we employ and adapt the LoFTR fine module~\cite{SunJ2021}. Instead of the pretrained ResNet feature maps of natural images at half resolution used by LoFTR~\cite{SunJ2021}, we incorporate our craquelure-specific feature maps at full resolution from which we extract $7 \times 7$ patches at the subpixel keypoint coordinates. The $7 \times 7$ patches are passed through a small, finetuned LoFTR module, that applies linear self- and cross-attention. 
Then, analogously to~\cite{WangQ2020,SunJ2021}, the center feature vector from modality 1 is correlated with the feature volume of modality 2. The output represents the matching probability of the pixels in modality 2 with the center vector in modality 1, and hence, the sub-pixel offset can be determined by softargmax.

\subsection{Training of the refinement modules}
The refinement modules are separately trained for both resolution levels (s0 and s1) using ground-truth sub-pixel keypoint pairs. The training of the refine module follows the description in~\cref{subsec_refinemodule}.
For the correlation modules, we further distinguish between the training of the craquelure feature volumes and the modified LoFTR fine module.

The craquelure feature volumes should describe the local areas around keypoints to find the best match between the two modalities. 
Viewed from this perspective, we can leverage descriptor learning via contrastive learning with a Triplet loss and online hard-negative mining.
For each keypoint pair, we define the anchor (descriptor of modality 1), the positive counterpart (correct descriptor in modality 2), and the negative one (hardest non-matching descriptor in modality 2). $32 \times 32 \times 3$ patches around the keypoints are fed through the frozen Craquelure backbone and the trainable feature fusion head, producing $32 \times 32 \times 128$ feature volumes. Then, the descriptors are extracted at sub-pixel level and triplet loss is computed.

To train our craquelure-specific adapted LoFTR fine module (CN-LofTR-f), we are setting up a different scenario. We assume that the best-matching position in the local neighborhood has already been identified in the previous step. The position should now be refined to a sub-pixel level. Therefore, we disturb the keypoint position of modality 2 with a small random uniform offset in the range of $[-1.5,1.5]$ pixels. The keypoints predicted by the correlation module and the sub-pixel ground-truth keypoints are minimized using MSE loss. All parts except for the LoFTR fine module are frozen.

\subsection{Outlier removal}
For non-rigid transformation estimation, accurate point correspondences are essential. To filter out remaining outliers, we compute a homography for each region pair using the robust USAC-MAGSAC++ estimator~\cite{RaguramR2013,BarathD2020} with a reprojection error threshold of 3 if there are at least 20 point pairs in the region. If we can predict a valid homography (according to \cref{subsec_filtering} (2)), points with a reprojection error larger than $th_{out}$ are discarded. Regions that could not be checked successfully in the first run are reinspected by merging them with neighboring regions (fully or partially) to compute the homography. This is repeated by neighborhood growing until all regions are successfully checked.

\subsection{TPS and homographic warping of huge images}
The images of the panel paintings at full XR resolution are too large to warp them in one piece. For backward warping, a displacement vector grid is computed, with x- and y-displacement values for each pixel in the output space. The displacement vector grid is divided into chunks. The bounding box of each region in the output space is mapped to the source image. Only the displacement vector chunks and the pixels of the corresponding source image regions are moved to the GPU for performing interpolation, and the output is transferred back to the CPU.

\section{Experiments and Results}

\subsection{Multi-modal Panel Painting Dataset Curation}
We created a novel multi-modal panel painting dataset with a high number of keypoint annotations.
The test set comprises large German panel paintings from the 13th to 16th century, provided by the Germanisches Nationalmuseum Nürnberg (GNM).
For the training and validation set, in addition to panel paintings, we also include 16th-century portraits by Lucas Cranach the Elder, already used in \cite{SindelA2021,SindelA2023}.

\subsubsection{Highly accurate keypoint annotations} 
For training of the CraquelureNet and the refine modules, we require highly accurate keypoint annotations, best set at the sub-pixel level. 
We increased the size of the keypoint detection and description dataset of \cite{SindelA2021,SindelA2023}. 
For keypoint detection and the refine module R, we annotated, for each modality \{VIS, IRR, UV, XR, MACRO\}, 20200 keypoints for training and 7494 for validation at the center of the craquelure and in the background. Background points were oversampled to match the number of craquelure points. For keypoint description, we annotated 13595 (training) and 7472 (validation) pairs of craquelure keypoints for each of the domains \{VIS-IRR, VIS-UV, VIS-XR/MACRO-XR, XR-IRR, UV-IRR\}. 
For the refine modules at level s0 and s1, we used subsets with XR and macro images.
In part, we increased the number of points by including different scales of the dataset.

For the evaluation, we created a large multi-modal test set comprising five domains, see \cref{tab-dataset-test}. XR-VIS and XR-IRR are our mixed-resolution datasets, since high-resolution XR is paired with lower-resolution VIS and IRR. 
Depending on the scaling factor between the VIS and XR resolutions, we divide the XR-VIS dataset into three subsets (Set1: 1.7-2.0, Set2: 3.0-3.8, Set3: 4.5-6.5). For Set1, we compute two scales (s0: XR 100\%, s1: VIS 100\%), and for Set2, Set3, and XR-IRR (XR/IRR: 2.5-3.2), we compute three scales (s0: XR 100\%, s1: XR 50\%, s2: VIS 100\% $|$ XR 25\% $|$ IRR 100\%).
The lowest scales (s1 of Set1 and s2 of the others), and the low-resolution VIS-IRR and VIS-UV, and the high-resolution XR-MACRO are used to test the one-stage method, while the higher scales are used for the step-by-step refinement methods.
The image sizes (mostly depicting only a part of the painting) range between $747 \times 1102$ and $7939 \times 42227$ pixels. 
We enhance the contrast of XR images by min-max normalization.
In average, we have annotated $122.4\pm83.5$ control points per image pair.

\subsubsection{Preregistered image patches}
The training of LightGlue requires images with ground-truth homographies. Generating synthetic counterparts to real images using CycleGANs~\cite{SindelA2022MICCAI}, did not yield good enough results for our multi-modal paintings. 
Thus, we preregistered the training and validation images using our pretrained CraquelureNet~\cite{SindelA2021} of cropped image pairs of roughly the same content area and scale. In case of non-rigid distortion of the cropped images, we already applied TPS otherwise homographies. We split the registered images into regions using Mean Shift clustering to extract square patches ($1472 \times 1472$) only in areas with keypoints.

\begin{table}[t]
\centering
\caption{Multi-modal Panel Dataset (Test)} 
\label{tab-dataset-test}
\tiny
\begin{tabular}{l rrr r rr r}
\toprule
Modalities & \multicolumn{3}{c}{VIS-XR} & XR-IRR & XR-MACRO & VIS-IRR & VIS-UV \\
& (Set1) & (Set2) & (Set3) & & & & \\
\midrule
Objects & 11 & 7 & 5 & 4 & 11 & 24 & 21 \\ 
Image Pairs & 20 & 18 & 20 & 18 & 68 & 60 & 21 \\ 
Point Pairs (Mean) & 193 & 153 & 161 & 107 & 75 & 124 & 149\\
\bottomrule 
\end{tabular}
\end{table}

\subsection{Implementation Details}

All methods are implemented in PyTorch and are trained using early stopping on the validation datasets. We use online data augmentation, \ie, color jittering, gamma adjustment, blur, sharpening, gaussian noise, flipping, and small rotation. 

CraquelureNet (CN) and the refinement modules are trained for $N_{epochs}$ using Adam solver and a linear learning rate decay to $0$ starting at epoch $N_{const}$ on an NVIDIA Titan Xp 12~GB GPU. For CN, we pretrain ResNet20 backbone + detection head (10.1M samples, $N_{epochs}=50$, $N_{const}=10$, learning rate $\eta=2\cdot10^{-3}$, batch size $BS=512$), and then jointly train ResNet20 backbone + detection head + description head (13.5M samples, $N_{epochs}=50$, $N_{const}=10$, $\eta=1\cdot10^{-3}$, $m=1$, $\lambda_\text{Det}=1$, $\lambda_\text{Desc}=1$,  $BS_{Det}=512$ and $BS_{Desc}=128$).  
We train three Craquelure Refine modules on different data subsets, \ie, R (5M samples, $N_{epochs}=50$, $N_{const}=5$), R-s0 and R-s1 (3.5M and 4.2M samples, $N_{epochs}=200$, $N_{const}=25$), 
and two Craquelure feature volumes (each 6.4M samples, $N_{epochs}=300$, $N_{const}=25$), and two CN-LoFTR-f modules (each 4.3M samples, $N_{epochs}=200$, $N_{const}=25$, using outdoor weights), all using ($\eta=4\cdot10^{-3}$, $BS=128$). 
For NCC, we use a Fast Fourier implementation~\cite{BertrandZNCC}.

For a fair comparison, we finetune LightGlue~\cite{LindenbergerP2023} (LG) to be used together with our methods CN or CN+R ($\eta=5.3\cdot10^{-4}$, $BS=112$), and SuperPoint~\cite{DeToneD2018} (SP) ($\eta=2\cdot10^{-4}$, $BS=104$) and Aliked~\cite{ZhaoX2023} ($\eta=1.6\cdot10^{-4}$, $BS=64$) as well as finetune LoFTR\cite{SunJ2021} ($\eta=1\cdot10^{-4}$, $BS=24$), all on our multi-modal dataset (1.5M samples, 100 epochs) using an image size of $1024 \times 1024$, the respective outdoor weights (for CN the weights of LG from SP), AdamW (weight decay of 0.001), and the One Cycle Learning Rate schedule.
LG is trained using mixed-precision on 2 NVIDIA A100 80~GB GPUs, and LoFTR using normal-precision on 8 NVIDIA A100 80~GB GPUs.
The applied batch size depends on the model size, and we scaled $\eta$ accordingly. 

We normalize all image data to \{-1,1\} for our methods. For the comparison methods, we use their own data normalization. 
We optimized hyperparameters for all methods on the validation datasets. 
For evaluation on our test datasets, we generally use the same setup for all methods unless otherwise stated.
Inference of all methods is performed on an NVIDIA Quadro RTX 8000 48~GB GPU.

\subsection{One-stage multi-modal non-rigid registration} \label{eval_one_stage}
In this experiment, we test our multi-modal non-rigid one-stage registration pipeline for the VIS-XR (Set 1 s1, Set 2-3 s2), XR-IRR (s2), VIS-IRR, VIS-UV, and XR-MACRO datasets. We evaluate the mean Euclidean error (ME) and maximum Euclidean error (MAE) of manual target control points and warped source control points after TPS estimation. The success rate (SR) is computed based on the error threshold~$\epsilon$. 

\subsubsection{Evaluation Protocol}
We evaluate our proposed method CN+R+LG against its ablations (CN+LG, CN+MNN), and state-of-the-art image matching methods, \ie, 
the fine-tuned models (SP+LG-P, Aliked+LG-P, LoFTR-P), the pretrained outdoor models (SP+LG-O, Aliked+LG-O, LoFTR-O, Roma-O~\cite{EdstedtJ2024}), and the pretrained multi-modal models (LoFTR-Mi~\cite{RenJ2025}, Roma-Mi~\cite{RenJ2025}, Roma-Ma~\cite{HeX2025}). 
We embed all methods into our patch-based registration pipeline. We used the validation set to optimize the parameters (patch size, patch stride) for the different groups: LG-based $(1536,1152)$, CN+MNN $(1024,768)$, LoFTR $(1024,768)$, Roma $(840,584)$.
Dependent on the patch size or due to hardware restrictions, we limit the number of keypoints per patch for the LG-based methods to $3840$, and for CN+MNN to $2560$, and restrict the number of matches per patch for LoFTR to $256$, and for Roma to $50$. 
For homography estimation, we tried both, weighted DLT (Kornia) and USAC+MAGSAC++ (OpenCV). DLT was more accurate, for the fine-tuned LG-based methods. For the dense methods, homography estimation only worked in combination with MAGSAC++. For CN+MNN we also use MAGSAC++.

\subsubsection{Results and Discussion}

\begin{figure*}
\scriptsize
\centering	
\includegraphics[width=0.8\textwidth]{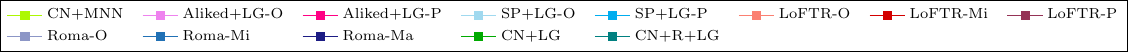}

\includegraphics[height=1.85cm]{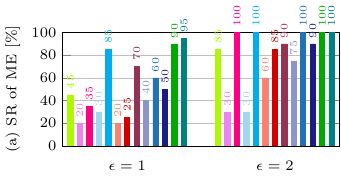}
\hfill
\includegraphics[height=1.85cm]{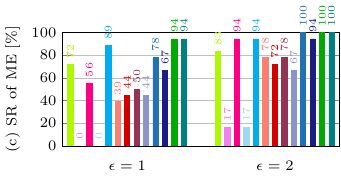}
\hfill
\includegraphics[height=1.85cm]{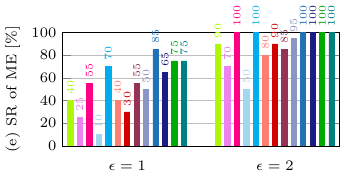}
\hfill
\includegraphics[height=1.85cm]{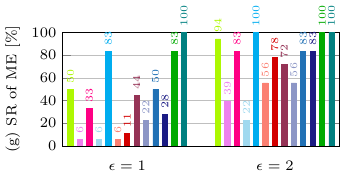}

\includegraphics[height=1.85cm]{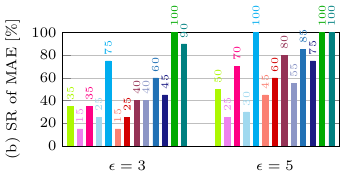}
\hfill
\includegraphics[height=1.85cm]{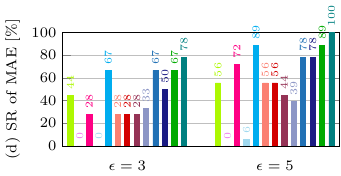}
\hfill
\includegraphics[height=1.85cm]{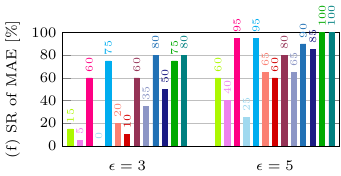}
\hfill
\includegraphics[height=1.85cm]{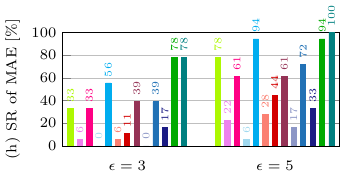}

\caption{Success rates of ME and MAE for VIS-XR and XR-IRR registration with error threshold $\epsilon$: (a,b) VIS-XR Set1 (s1), (c,d) VIS-XR Set2 (s2), (e,f) VIS-XR Set3 (s2), (g,h) XR-IRR (s2). VFC is applied as post-processing. Methods with '-P' are fine-tuned on our Multi-modal Panel Painting dataset.}
\label{fig-test-01}
\end{figure*}

\begin{table*}[t]
\centering
\caption{Quantitative evaluation for VIS-XR and XR-IRR. VFC is applied as post-processing. Methods with `-P' are fine-tuned on our Multi-modal Panel Painting dataset. Mean ME and MAE are marked as `--' if not all images could be registered. Best results are highlighted in bold.} 
\label{tab-test-01}
\tiny
\begin{tabular}{lrr@{\hspace{8pt}}crr@{\hspace{8pt}}crr@{\hspace{8pt}}crr}
\toprule
& \multicolumn{8}{c}{VIS-XR (lowest resolution level)} && \multicolumn{2}{c}{XR-IRR (lowest resolution level)}\\
\cmidrule{2-9}
\cmidrule{11-12}
 & \multicolumn{2}{c}{Set1 (scale 1.7-2.0)} && \multicolumn{2}{c}{Set2 (scale 3.0-3.8)} && \multicolumn{2}{c}{Set3 (scale 4.5-6.5)} && \multicolumn{2}{c}{(scale 2.5-3.2)} \\
\cmidrule{2-3}
\cmidrule{5-6}
\cmidrule{8-9}
\cmidrule{11-12}
Method & ME $\downarrow$ & MAE $\downarrow$ && ME $\downarrow$ & MAE $\downarrow$ && ME $\downarrow$ & MAE $\downarrow$ && ME $\downarrow$ & MAE $\downarrow$ \\
& Mean$\pm$Std & Mean$\pm$Std && Mean$\pm$Std & Mean$\pm$Std && Mean$\pm$Std & Mean$\pm$Std && Mean$\pm$Std & Mean$\pm$Std \\
\midrule
Aliked+LG-P + DLT & 1.18$\pm$0.39 & 3.89$\pm$1.90 & & 1.11$\pm$0.43 & 4.60$\pm$2.56 & & 0.98$\pm$0.26 & 2.87$\pm$0.82 & & 1.37$\pm$0.74 & 5.36$\pm$3.25 \\
SP+LG-P + DLT & 0.81$\pm$0.17 & 2.48$\pm$0.71 & & 0.77$\pm$0.39 & 3.88$\pm$5.23 & & 0.89$\pm$0.22 & 2.86$\pm$0.98 & & 0.80$\pm$0.17 & 3.16$\pm$0.99 \\
CN+LG + DLT & 0.81$\pm$0.16 & \textbf{2.26$\pm$0.57} & & 0.69$\pm$0.14 & 2.96$\pm$2.11 & & 0.88$\pm$0.22 & 2.71$\pm$0.59 & & 0.79$\pm$0.13 & 2.83$\pm$0.95 \\
CN+R+LG + DLT & \textbf{0.80$\pm$0.16} & 2.35$\pm$0.74 & & \textbf{0.64$\pm$0.12} & \textbf{2.37$\pm$0.90} & & 0.85$\pm$0.23 & \textbf{2.51$\pm$0.72} & & \textbf{0.74$\pm$0.12} & \textbf{2.65$\pm$0.86} \\
\midrule
LoFTR-O + Magsac & 3.00$\pm$3.84 & 11.45$\pm$14.43 & & 4.70$\pm$13.59 & 18.32$\pm$52.95 & & 1.89$\pm$2.27 & 6.83$\pm$7.04 & & \multicolumn{1}{c}{--} & \multicolumn{1}{c}{--}\\
LoFTR-Mi + Magsac & 1.59$\pm$1.11 & 6.44$\pm$6.14 & & 2.84$\pm$3.71 & 12.49$\pm$16.37 & & 1.52$\pm$0.83 & 6.09$\pm$3.76 & & 4.25$\pm$11.41 & 15.25$\pm$35.26 \\
LoFTR-P + Magsac & 9.96$\pm$39.85 & 43.04$\pm$175.12 & & 32.97$\pm$87.98 & 206.09$\pm$447.06 & & 25.37$\pm$104.05 & 47.91$\pm$173.23 & & 31.02$\pm$81.21 & 232.69$\pm$571.72 \\
Roma-O + Magsac & 1.85$\pm$1.96 & 6.75$\pm$7.13 & & 1.65$\pm$1.21 & 7.96$\pm$7.08 & & 1.10$\pm$0.45 & 4.99$\pm$3.98 & & 7.05$\pm$19.62 & 27.25$\pm$65.55 \\
Roma-Mi + Magsac & 0.89$\pm$0.28 & 3.38$\pm$2.33 & & 0.87$\pm$0.50 & 5.13$\pm$6.30 & & \textbf{0.83$\pm$0.27} & 2.88$\pm$1.68 & & 2.47$\pm$4.68 & 9.56$\pm$18.25 \\
Roma-Ma + Magsac & 1.13$\pm$0.51 & 4.44$\pm$3.39 & & 0.97$\pm$0.51 & 5.46$\pm$6.32 & & 0.95$\pm$0.36 & 4.00$\pm$2.73 & & 68.36$\pm$283.14 & 347.58$\pm$1443.90 \\
\bottomrule
\end{tabular}
\end{table*}

\begin{figure*}
\scriptsize
\centering
\subfloat[][VIS\label{fig-test-qual-01a}]{
\includegraphics[height=3.7cm]{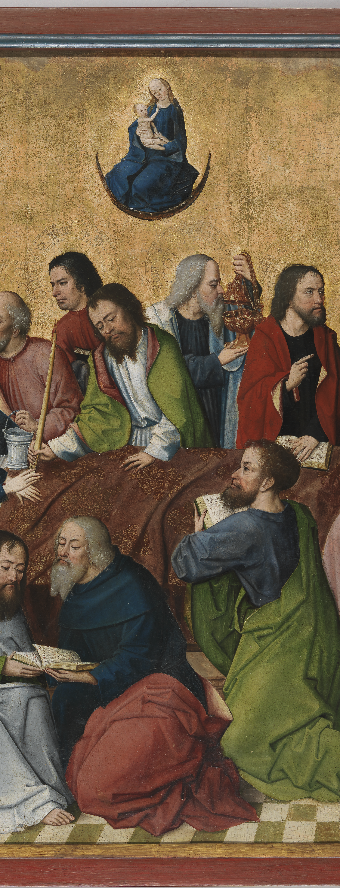}
}
\subfloat[][XR\label{fig-test-qual-01b}]{
\includegraphics[height=3.7cm]{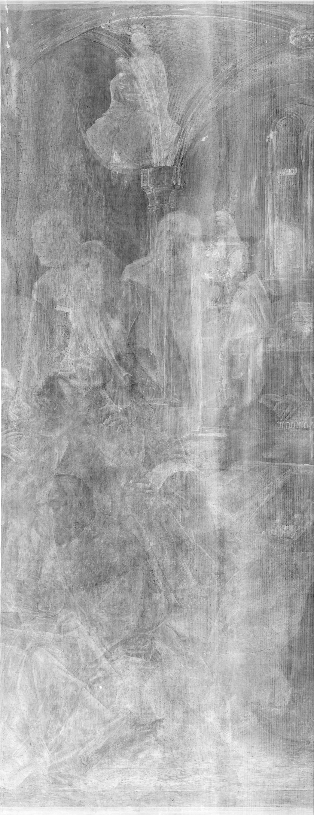}
}
\subfloat[][LoFTR-P\label{fig-test-qual-01c}]{
\includegraphics[height=3.7cm]{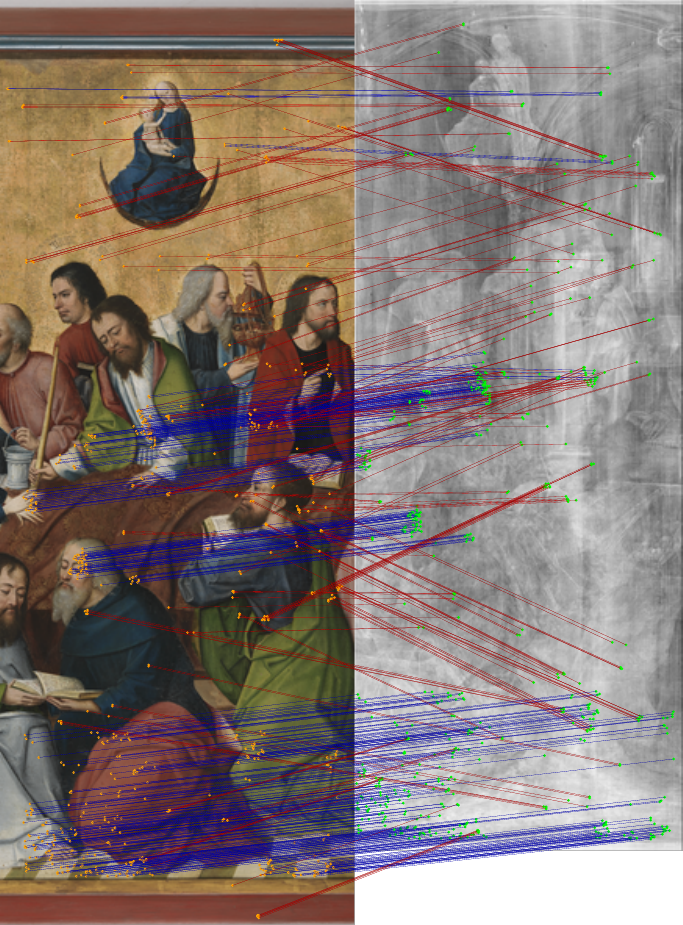}
}
\subfloat[][Roma-Mi\label{fig-test-qual-01d}]{
\includegraphics[height=3.7cm]{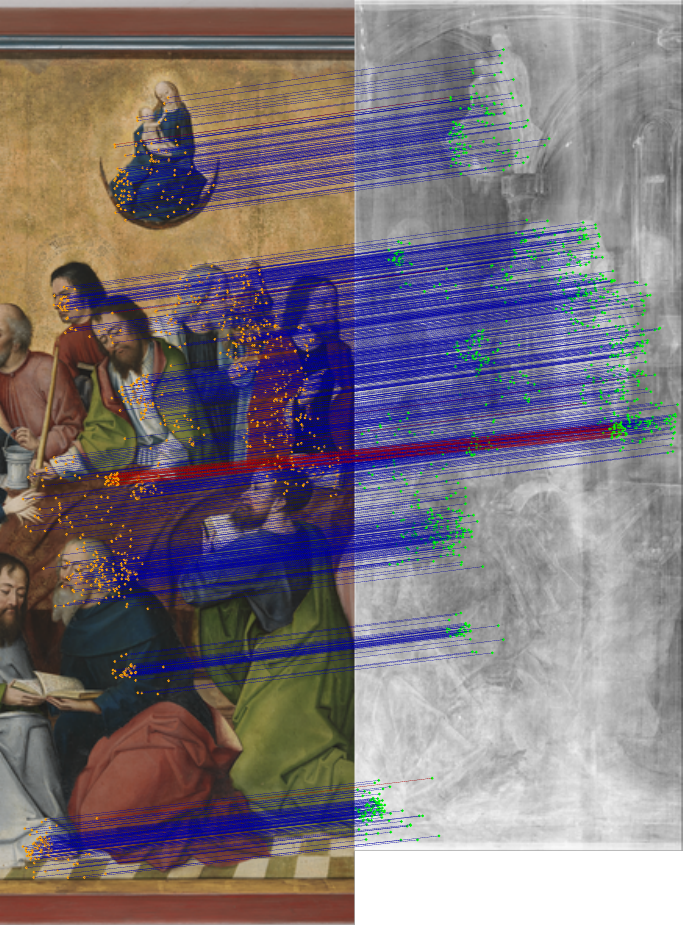}
}
\subfloat[][Aliked+LG-P\label{fig-test-qual-01e}]{
\includegraphics[height=3.7cm]{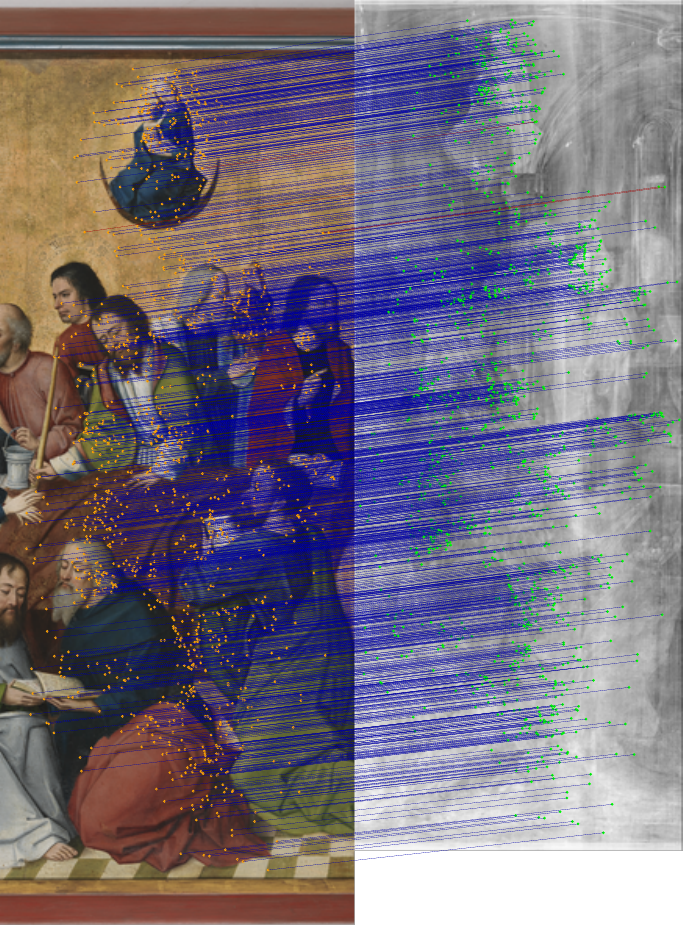}
}
\subfloat[][SP+LG-P\label{fig-test-qual-01f}]{
\includegraphics[height=3.7cm]{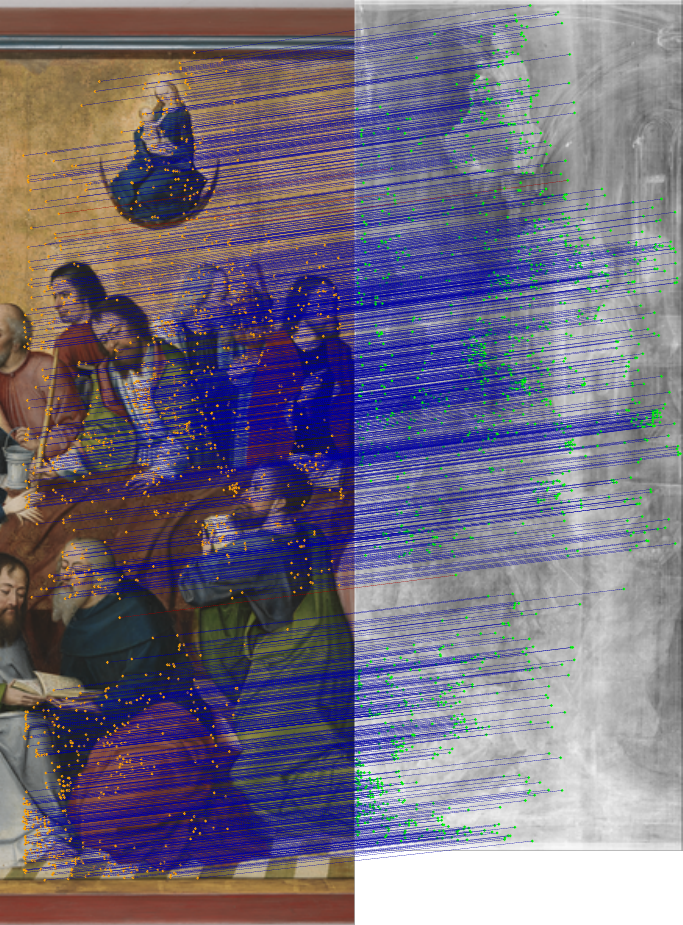}
}
\subfloat[][CN+R+LG\label{fig-test-qual-01g}]{
\includegraphics[height=3.7cm]{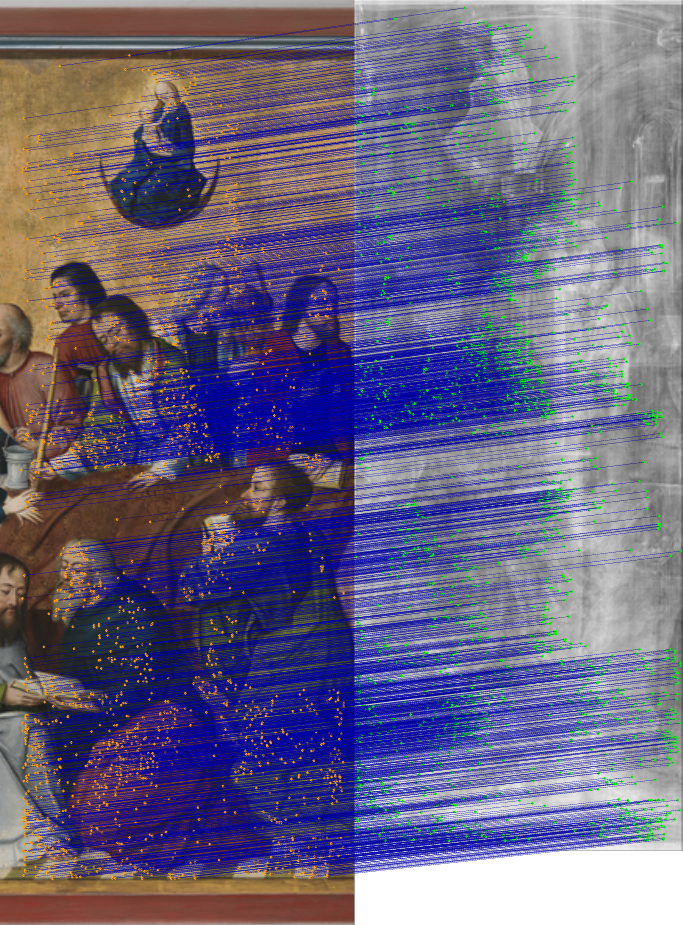}
}
\caption{Qualitative results for one example of VIS-XR Set2 (s2) registration. In (c-g), the keypoint correspondences (target: orange, source: green) and their connections (blue: kept by VFC, red: filtered out by VFC) are shown. Best viewed when zooming in. Image sources: Meister des Marienlebens, Tod Mariae (Innenseite), VIS and XR, Germanisches Nationalmuseum, Nuremberg, on loan from Wittelbacher Ausgleichsfond, Bayerische Staatsgemäldesammlungen, Gm 20, all rights reserved}
\label{fig-test-qual-01}
\end{figure*}

\begin{table*}[t]
\centering
\caption{Quantitative evaluation for XR-MACRO, VIS-IRR, and VIS-UV. VFC is applied as post-processing. Methods with `-P' are fine-tuned on our Multi-modal Panel Painting dataset. Mean ME and MAE are marked as `--' if not all images could be registered. Best results are highlighted in bold.}
\label{tab-test-02}
\tiny
\begin{tabular}{l r@{\hspace{5pt}}r@{\hspace{5pt}} r@{\hspace{5pt}}r@{\hspace{5pt}}r@{\hspace{5pt}}r c@{} r@{\hspace{5pt}}r@{\hspace{5pt}}r@{\hspace{5pt}}r@{\hspace{5pt}}r@{\hspace{5pt}}r c@{} r@{\hspace{5pt}}r@{\hspace{5pt}}r@{\hspace{5pt}}r@{\hspace{5pt}}r@{\hspace{5pt}}r}
\toprule
 & \multicolumn{6}{c}{XR-MACRO} && \multicolumn{6}{c}{VIS-IRR} && \multicolumn{6}{c}{VIS-UV} \\
\cmidrule{2-7}
\cmidrule{9-14} 
\cmidrule{16-21}
 & \multicolumn{2}{c}{SR-ME $\uparrow$} & \multicolumn{2}{c}{SR-MAE $\uparrow$} & ME $\downarrow$ & MAE $\downarrow$ && \multicolumn{2}{c}{SR-ME $\uparrow$} & \multicolumn{2}{c}{SR-MAE $\uparrow$} & ME $\downarrow$ & MAE $\downarrow$ && \multicolumn{2}{c}{SR-ME $\uparrow$} & \multicolumn{2}{c}{SR-MAE $\uparrow$} & ME $\downarrow$ & MAE $\downarrow$ \\
& $\epsilon=1$ & $\epsilon=2$ & $\epsilon=3$ & $\epsilon=5$ & Mean$\pm$Std & Mean$\pm$Std && $\epsilon=1$ & $\epsilon=2$ & $\epsilon=3$ & $\epsilon=5$ & Mean$\pm$Std & Mean$\pm$Std && $\epsilon=1$ & $\epsilon=2$ & $\epsilon=3$ & $\epsilon=5$ & Mean$\pm$Std & Mean$\pm$Std \\
\midrule
CN+MNN + Magsac & 58.8 & 95.6 & 57.4 & 82.4 & 1.05$\pm$0.50 & 3.56$\pm$2.65 & & 11.7 & \textbf{100.0} & 31.7 & 96.7 & 1.27$\pm$0.25 & 3.47$\pm$1.07 & & 4.8 & 81.0 & 9.5 & 66.7 & 1.99$\pm$1.16 & 6.40$\pm$5.47 \\
\midrule
Aliked+LG-O + DLT & 19.1 & 64.7 & 19.1 & 44.1 & \multicolumn{1}{c}{--} & \multicolumn{1}{c}{--} & & 6.7 & 96.7 & 31.7 & 85.0 & 1.37$\pm$0.36 & 3.93$\pm$1.74 & & 0.0 & 90.5 & 4.8 & 71.4 & 1.76$\pm$0.55 & 4.69$\pm$1.67 \\
Aliked+LG-P + DLT & 48.5 & 97.1 & 55.9 & 92.6 & 1.10$\pm$0.68 & 3.20$\pm$2.17 & & 10.0 & \textbf{100.0} & 28.3 & 96.7 & 1.29$\pm$0.23 & 3.55$\pm$0.97 & & 0.0 & 90.5 & 0.0 & 85.7 & 1.59$\pm$0.30 & 4.15$\pm$0.77 \\
SP+LG-O + DLT & 41.2 & 73.5 & 33.8 & 54.4 & \multicolumn{1}{c}{--} & \multicolumn{1}{c}{--} & & 13.3 & 98.3 & 33.3 & 93.3 & 1.29$\pm$0.27 & 3.59$\pm$1.18 & & 4.8 & 90.5 & 9.5 & 76.2 & 1.59$\pm$0.30 & 4.24$\pm$1.07 \\
SP+LG-P + DLT & 67.6 & 98.5 & 73.5 & 97.1 & 0.98$\pm$0.96 & 2.82$\pm$2.41 & & 15.0 & \textbf{100.0} & 36.7 & 98.3 & 1.25$\pm$0.23 & 3.33$\pm$0.77 & & 4.8 & 95.2 & 9.5 & \textbf{100.0} & 1.55$\pm$0.29 & 3.92$\pm$0.60 \\
CN+LG + DLT & 75.0 & \textbf{100.0} & \textbf{79.4} & 98.5 & 0.89$\pm$0.23 & 2.50$\pm$0.80 & & 16.7 & \textbf{100.0} & 35.0 & \textbf{100.0} & 1.25$\pm$0.24 & \textbf{3.27$\pm$0.69} & & 4.8 & \textbf{100.0} & 4.8 & 90.5 & 1.56$\pm$0.28 & 3.94$\pm$0.69 \\
CN+R+LG + DLT & \textbf{76.5} & \textbf{100.0} & 77.9 & \textbf{100.0} & \textbf{0.85$\pm$0.22} & \textbf{2.44$\pm$0.72} & & 16.7 & \textbf{100.0} & 33.3 & \textbf{100.0} & 1.24$\pm$0.24 & \textbf{3.27$\pm$0.70} & & 4.8 & \textbf{100.0} & 4.8 & \textbf{100.0} & 1.55$\pm$0.29 & \textbf{3.84$\pm$0.59} \\
\midrule
LoFTR-O + Magsac & 29.4 & 89.7 & 39.7 & 75.0 & 1.68$\pm$2.10 & 5.11$\pm$5.83 & & 8.3 & \textbf{100.0} & 26.7 & 95.0 & 1.33$\pm$0.25 & 3.59$\pm$0.78 & & 0.0 & 66.7 & 0.0 & 61.9 & 1.78$\pm$0.38 & 4.94$\pm$1.51 \\
LoFTR-Mi + Magsac & 22.1 & 88.2 & 41.2 & 72.1 & \multicolumn{1}{c}{--} & \multicolumn{1}{c}{--} & & 3.3 & \textbf{100.0} & 15.0 & 86.7 & 1.41$\pm$0.24 & 3.92$\pm$1.07 & & 0.0 & 61.9 & 4.8 & 57.1 & 1.95$\pm$0.63 & 5.86$\pm$2.95 \\
LoFTR-P + Magsac & 36.8 & 73.5 & 38.2 & 69.1 & 870$\pm$6190 & 7319$\pm$49952 & & 13.3 & \textbf{100.0} & 38.3 & 98.3 & 1.25$\pm$0.23 & 3.36$\pm$0.74 & & 4.8 & 90.5 & 9.5 & 76.2 & 1.61$\pm$0.30 & 4.20$\pm$0.87 \\
Roma-O + Magsac & 51.5 & 95.6 & 47.1 & 75.0 & 1.16$\pm$0.91 & 3.99$\pm$3.38 & & 15.0 & \textbf{100.0} & 35.0 & 96.7 & 1.24$\pm$0.24 & 3.36$\pm$0.87 & & 4.8 & 95.2 & 9.5 & 95.2 & \textbf{1.54$\pm$0.29} & 3.87$\pm$0.61 \\
Roma-Mi + Magsac & 73.5 & \textbf{100.0} & 75.0 & 97.1 & 0.86$\pm$0.25 & 2.55$\pm$1.16 & & 15.0 & \textbf{100.0} & 38.3 & 96.7 & \textbf{1.23$\pm$0.24} & 3.30$\pm$0.79 & & 4.8 & 90.5 & 9.5 & 95.2 & 1.70$\pm$0.82 & 4.32$\pm$2.41 \\
Roma-Ma + Magsac & 61.8 & 97.1 & 58.8 & 86.8 & 0.99$\pm$0.45 & 3.19$\pm$2.10 & & 16.7 & \textbf{100.0} & 35.0 & 98.3 & 1.24$\pm$0.24 & 3.34$\pm$0.84 & & 4.8 & 95.2 & 9.5 & 85.7 & 1.71$\pm$0.83 & 4.57$\pm$2.73 \\
\bottomrule
\end{tabular}
\end{table*}

\begin{figure*}
\scriptsize
\centering
\subfloat[][MACRO\label{fig-test-qual-02a}]{
\includegraphics[height=3.4cm]{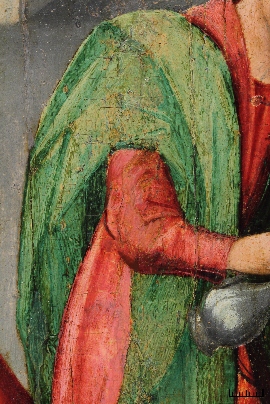}
}
\subfloat[][XR\label{fig-test-qual-02b}]{
\includegraphics[height=3.4cm]{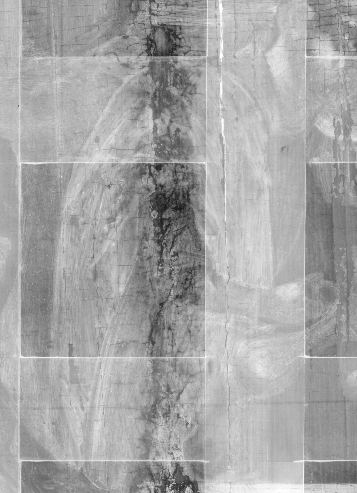}
}
\subfloat[][LoFTR-P\label{fig-test-qual-02c}]{
\includegraphics[height=3.4cm]{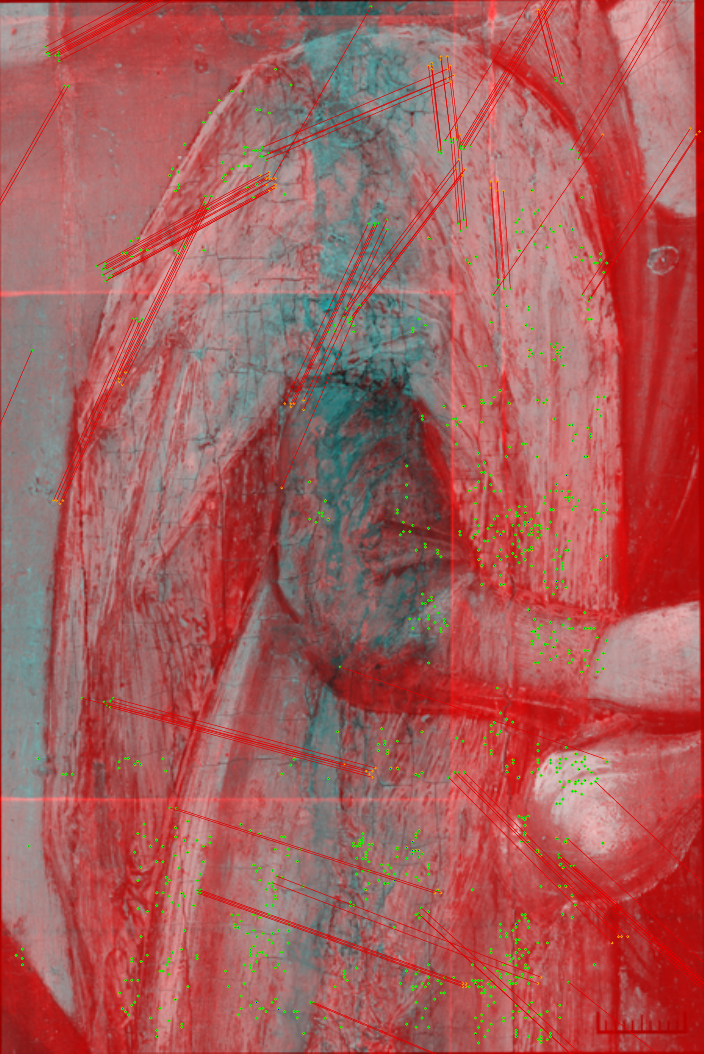}
}
\subfloat[][Roma-Mi\label{fig-test-qual-02d}]{
\includegraphics[height=3.4cm]{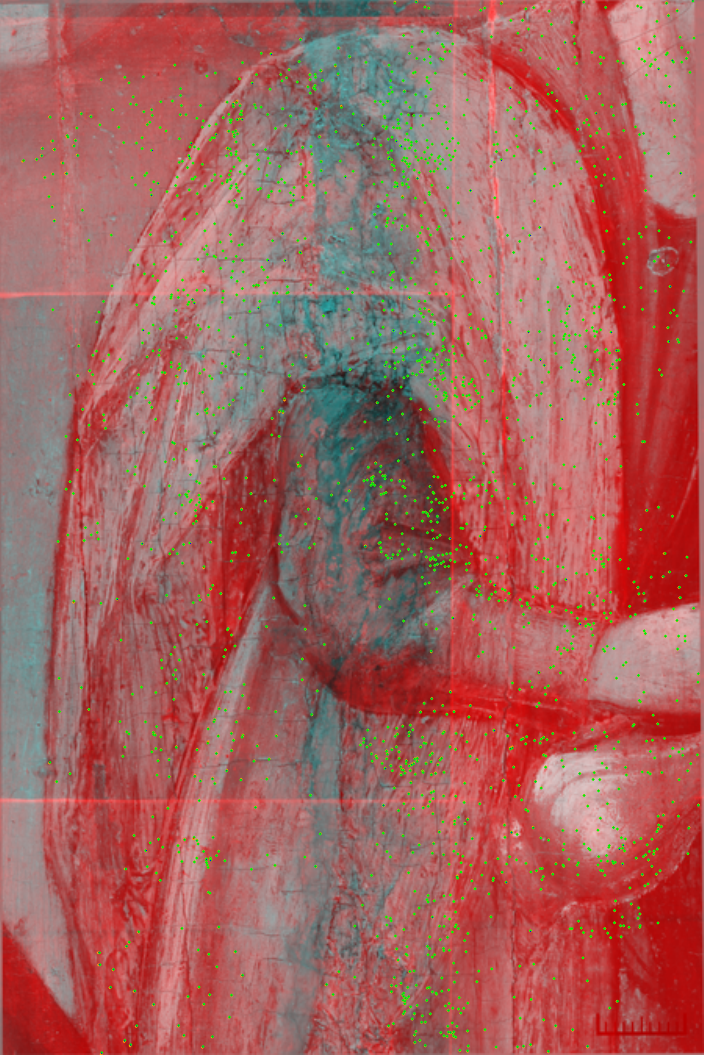}
}
\subfloat[][Aliked+LG-P\label{fig-test-qual-02e}]{
\includegraphics[height=3.4cm]{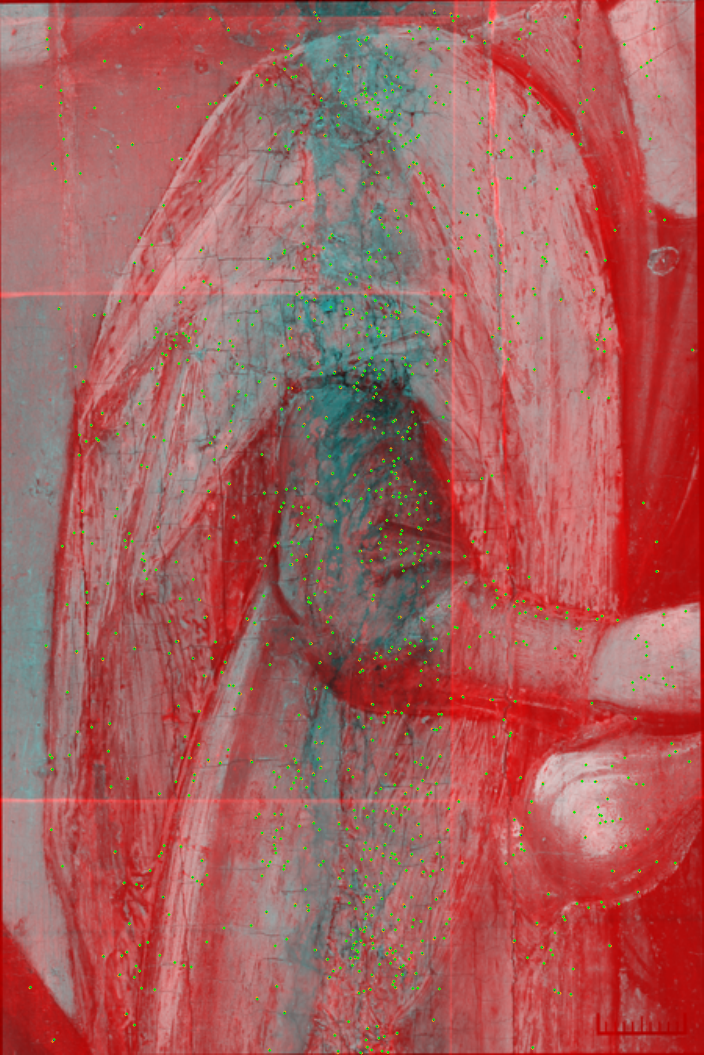}
}
\subfloat[][SP+LG-P\label{fig-test-qual-02f}]{
\includegraphics[height=3.4cm]{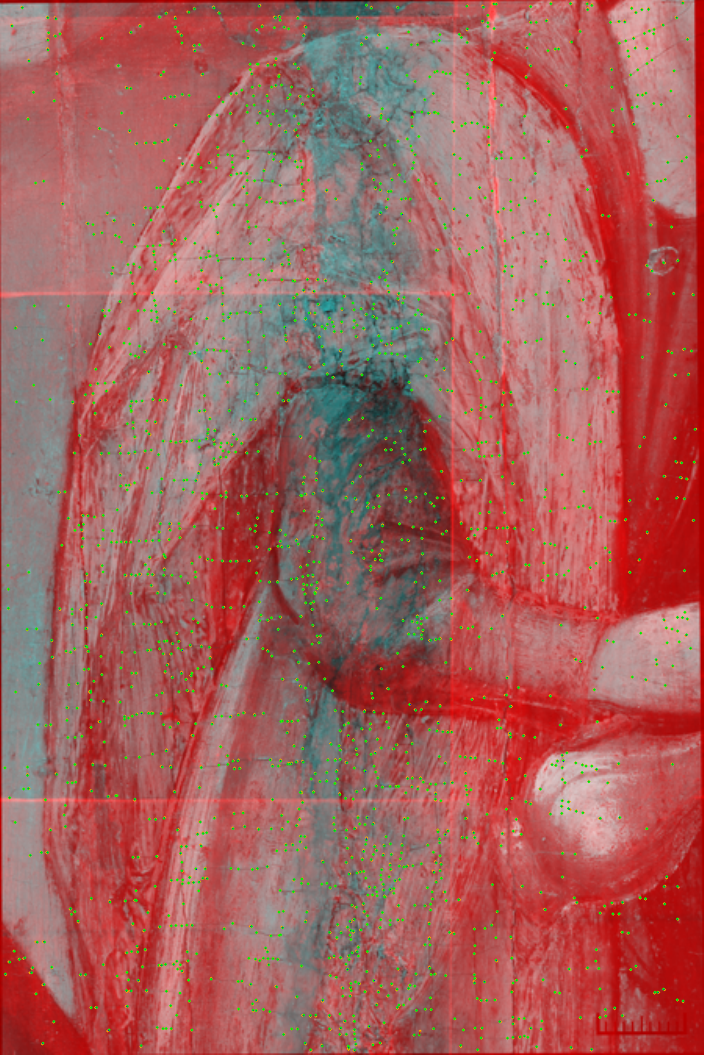}
}
\subfloat[][CN+R+LG\label{fig-test-qual-02g}]{
\includegraphics[height=3.4cm]{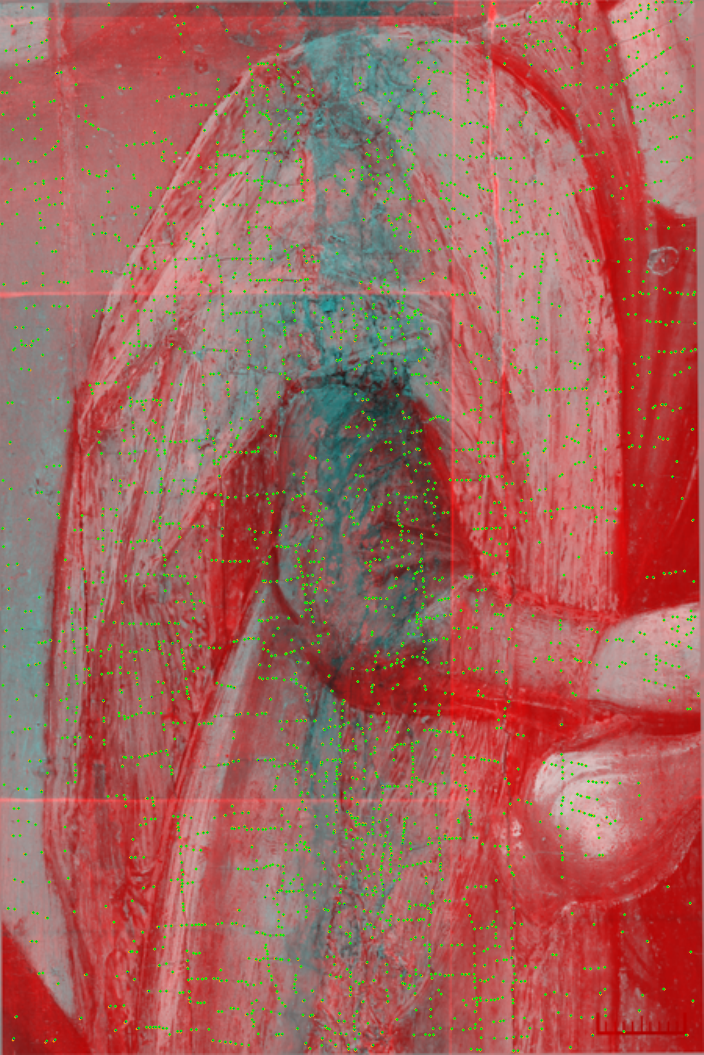}
}
\caption{Qualitative results for one example of XR-MACRO registration. In (c-g), the false color overlays w.r.t. the global homography with superimposed registered keypoints (target: orange, warped source: green) and displacement vectors (blue: kept by VFC, red: filtered out by VFC) are shown. Image sources: Gabriel Mälesskircher, Maria Magdalena salbt die Füße Christi, MACRO and XR, Germanisches Nationalmuseum, Nuremberg, Gm 1463, all rights reserved}
\label{fig-test-qual-02}
\end{figure*}

The quantitative results for VIS-XR and XR-IRR registration are summarized in \cref{fig-test-01} and \cref{tab-test-01} and for XR-MACRO, VIS-IRR, and VIS-UV registration in \cref{tab-test-02}. 
CN+R+LG outperforms the competing methods across all datasets, achieving a success rate of 100\% for ME $<$ 2 pixels and MAE $<$ 5 pixels, and for XR-IRR, even for ME $<$ 1 pixel. 
For the VIS-XR, XR-IRR, and XR-MACRO datasets, we obtain, for our methods CN+R+LG and CN+LG, mean MEs well below 1 pixel and mean MAs below 3 pixels. SP+LG-P also performs well, but with overall worse results. Roma-Mi as well achieves mean ME values below 1 pixel for VIS-XR and XR-MACRO, but higher mean MAE values, and does not work well for XR-IRR.
For VIS-IRR and VIS-UV, the mean ME and MAE values of all methods are generally a bit higher; however, for VIS-IRR, almost all methods except for the sparse pretrained models achieve 100\% SR-ME at $\epsilon=2$. 

Overall, fine-tuning LG for our CraquelureNet yields a huge performance boost compared to using the MNN matcher. In addition, adding the refine module further improves success rates and mean error statistics, except for VIS-XR Set1, where both configurations are comparable. 
Fine-tuning LG also proved beneficial for SP and Aliked, since their pretrained outdoor models have difficulties with all XR datasets. 
We see the same trend for fine-tuning LoFTR. LoFTR-P shows higher success rates than the original LoFTR-O and multi-modal LoFTR-Mi.
For the Roma variants, Roma-Mi overall performs better than Roma-O and Roma-Ma.

Dense methods, such as Roma and LoFTR, achieve the best accuracy on computer vision datasets due to their dense point distribution, but they reach their limits with very large images, both computationally and in terms of accuracy. The inference time to perform the one-stage registration for sparse methods (CN+R+LG $32\,s$, SP+LG-P $23\,s$, and Aliked+LG-P $20\,s$ per image for VIS-XR Set2 (s2)) is considerably shorter than for the dense methods (Roma $115\,s$ and LoFTR $40\,s$ per image). Thus, our method is fast and yields the highest accuracy for registration of panel paintings.

The \cref{fig-test-qual-01,fig-test-qual-02} show qualitative results for VIS-XR Set2 and for XR-MACRO registration for the best performing methods of each group. 
For each method, the matched keypoint correspondences are visualized in \cref{fig-test-qual-01}, and the false-color overlay after global homography computation with superimposed keypoints and displacement vectors in \cref{fig-test-qual-02}. Blue-colored matches and displacement vectors are used for TPS computation, while VFC has filtered out the red ones.
Especially, in the setting of coarse-to-fine registration, a balanced distribution of keypoints is essential to ensure sufficient precision in displacement estimation across all regions for TPS computation while still limiting the number of TPS control points. 
By applying the dense methods patch-wise, we observe clumps of point accumulations and also very sparse areas with hardly any points. Conversely, our craquelure-based method distributes keypoints equally across crack locations. \mbox{SP+LG-P} and Aliked+LG-P also partly place keypoints on cracks; however, the keypoints are less balanced and distributed. 
The image of \cref{fig-test-qual-01} poses another difficulty, as the panel has been painted on both sides. The XR image penetrates the complete panel; hence, the intensity values represent an addition of both paint layers.  Consequently, a mix of motifs and cracks from the front and back of the panel is visible in XR, but only from one side of the panel in VIS.  

\subsection{Coarse-to-fine multi-modal non-rigid registration}
In this experiment, we evaluate different refinement approaches for XR-VIS and XR-IRR registration. We conduct an ablation study for the modules of our proposed refinement method regarding a subset of ground-truth control point positions and measure the registration performance of the final TPS estimation, comparing two approaches. 

\subsubsection{Evaluation Protocol}
For the ablation study, we select a subset of ground-truth control points whose target points (XR) are each within a 2-pixel radius of a keypoint detected by CN+R+LG on VIS-XR (Set1 s1, Set2-3 s2) and XR-IRR (s2). We perform the refinement for all keypoints without outlier removal. This enables direct comparison of upscaled, refined, and ground-truth points of the chosen common subset. 

For the registration performance evaluation (SR, ME, MAE), we rerun the refinement and perform outlier removal after the last refinement level. For this experiment, we compare our refinement method based on CN keypoints with an alternative approach using SP keypoints and a modified KeyPt2SubPx module.
KeyPt2SubPx~\cite{KimS2024} requires the descriptors and patches extracted at the keypoints from the input images and from the keypoint score maps.
For the step-by-step refinement, we want to apply the refinement to the upscaled keypoints in each level; however, KeyPt2SubPx is not directly applicable in this case.
Our modified SP-KeyPt2SubPx module first extracts a dense score map and a coarse descriptor grid for the input patch of the current level using SP. From these, patches with a radius of 5 pixels can be extracted, and descriptors can be interpolated at the upscaled keypoint positions. The refined keypoints are then obtained with the pretrained KeyPt2SubPx. With this, we can plug SP-KeyPt2SubPx into our refinement pipeline.
For the first level, the region size is $768$ and for the second level $1536$. 

\subsubsection{Results and Discussion}

\begin{figure}
\scriptsize
\centering
\subfloat[\label{fig-test-ref-01a}]{
\begin{minipage}[c]{.36\textwidth}
    \centering
	\includegraphics[height=0.3cm]{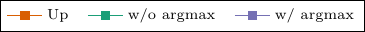}	
	\includegraphics[width=\textwidth]{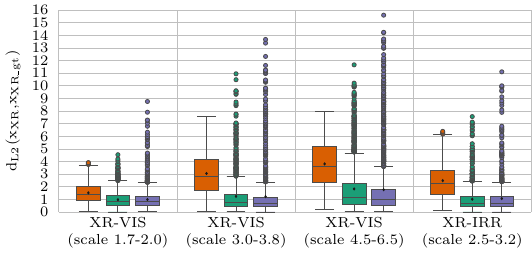}
\end{minipage}
}

\subfloat[\label{fig-test-ref-01b}]{
\begin{minipage}[c]{.47\textwidth}
	\centering
	\includegraphics[width=\textwidth]{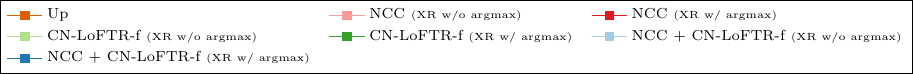}
	\includegraphics[width=.8\textwidth]{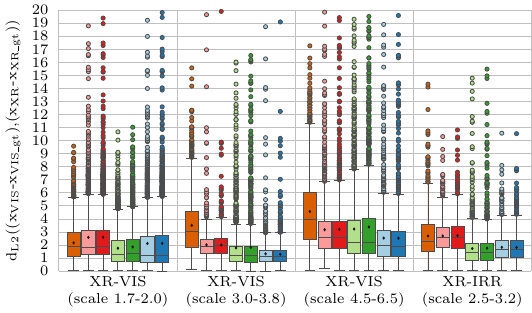}
\end{minipage}
}

\caption{Ablation study for XR-VIS and XR-IRR refinement. In (a), XR refinement reduces the distances to ground-truth (GT) points compared to upscaling. In (b), small distances between the vector differences to the GT points of both modalities indicate high consistency between the keypoint pairs.}
\label{fig-test-ref-01}
\end{figure}

\begin{figure}
\scriptsize
\centering
\begin{minipage}[t]{.08\textwidth}
	\centering
	\includegraphics[height=0.7cm]{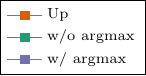}
\end{minipage}
\hspace{0.1cm}
\subfloat[\label{fig-test-ref-02a}]{
\includegraphics[width=0.15\textwidth]{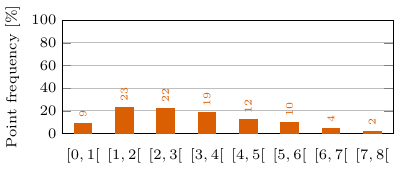}
}
\hspace{0.1cm}
\subfloat[\label{fig-test-ref-02b}]{
\includegraphics[width=0.15\textwidth]{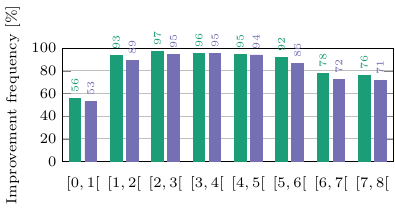}
}

\subfloat[\label{fig-test-ref-02c}]{
\includegraphics[width=0.47\textwidth]{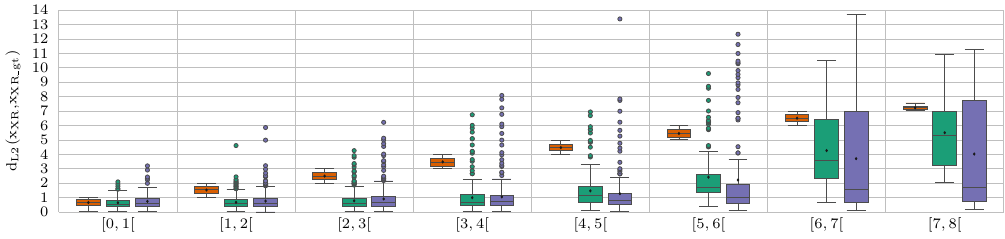}
}

\caption{Ablation study for XR refinement (XR-VIS Set2 s0) regarding different groups for the initial ground-truth (GT) displacements of the upscaled points. In (a), the point frequency, in (b), the improvement frequency (refinement over upscaling), and in (c), the GT distances (w/ argmax best median) are plotted for all pixel distance groups.}
\label{fig-test-ref-02}
\end{figure}

\begin{figure}
\scriptsize
\centering
\begin{minipage}[c]{.47\textwidth}
	\centering
	\includegraphics[width=\textwidth]{graphics/test-refine-gt/legend-vis-half}
\end{minipage}

\subfloat[\label{fig-test-ref-03a}]{
	\includegraphics[width=0.47\textwidth]{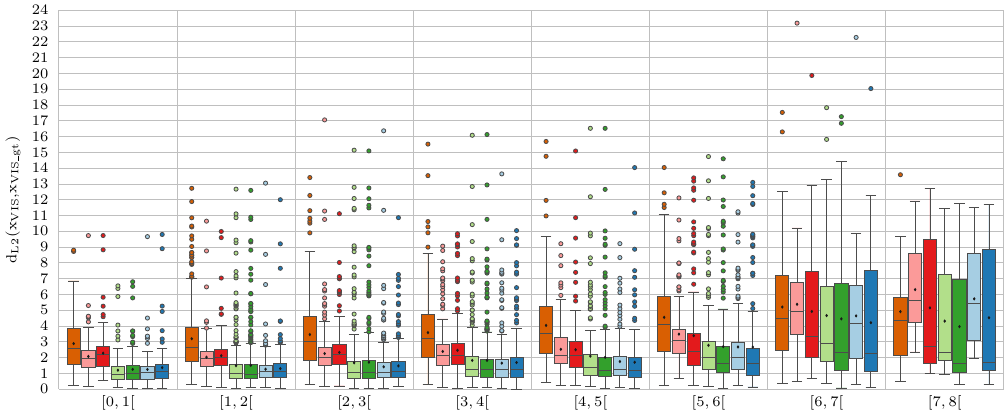}
}

\subfloat[\label{fig-test-ref-03b}]{
	\includegraphics[width=0.47\textwidth]{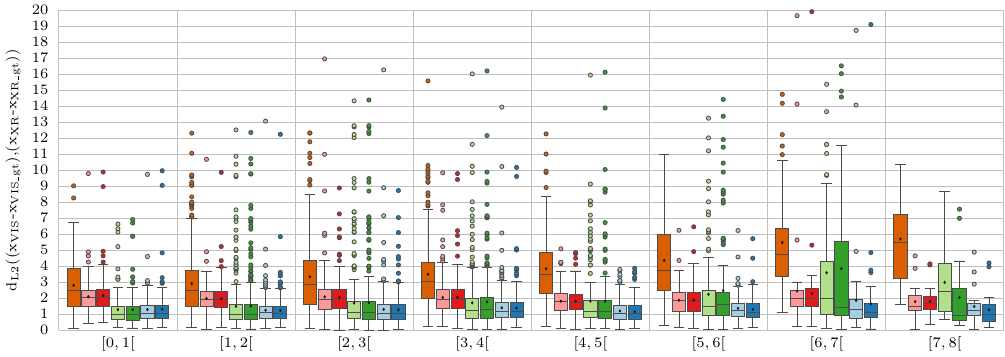}
}

\caption{Ablation study for VIS refinement (XR-VIS Set2 s0) regarding the GT distance groups of \cref{fig-test-ref-02}. In (a), the GT distances and in (b), the distances between the vector differences to the GT points of both modalities are plotted. }
\label{fig-test-ref-03}
\end{figure}

The quantitative results of the ablation study for our keypoint refinement method on all mixed-resolution datasets are shown in \cref{fig-test-ref-01}. First, we compare the refined and the upscaled XR keypoints against the ground-truth XR control points. The box plots show an approximation of the refined keypoints to the XR control points. For XR-VIS (Set1) and XR-IRR, the refinement without argmax results in a slightly lower mean (diamond in the boxplot) and fewer outliers. For XR-VIS (Set2-3), the refinement with argmax reduces the distances to the ground truth more strongly (smaller mean, median, and interquartile range (IQR)).
The refinement of the keypoints in the second modality (VIS and IRR) depends on the keypoints in the first modality (XR). 
Keypoints in the second modality should be shifted to best match the refined keypoint position in the first modality. 
Hence, an improvement over the upscaled points can be measured when the distance between the vector differences of the refined points and the ground truth points is reduced. The ablation study shows results for the refinement variants of modality 1 paired with the ones of modality 2 (NCC, CN-LoFTR-f, both). For the datasets with a smaller scaling factor difference (XR-VIS Set1 and XR-IRR), the combination of XR without argmax and CN-LoFTR-f has the lowest mean and IQR. For the other two datasets with a larger scaling factor difference (XR-VIS (Set2-3)), the complete refinement configuration (XR with argmax, NCC + CN-LoFTR-f) results in the lowest mean, median, and IQR. 

For XR-VIS (Set2), we further analyze the refinement results in \cref{fig-test-ref-02} with respect to the initial displacement between upscaled XR keypoints and ground-truth control points. \cref{fig-test-ref-02a}, shows the point frequency in \% of the upscaled XR points divided into eight bins with ground-truth distances between 0 and 8 pixels. The most upscaled points are between 1 and 4 pixels. For these bins, we compared each refined keypoint to the upscaled keypoint and computed the percentage of points where the ground-truth distance was reduced by the refinement relative to the upscaled point. Except for the first bin (0 to 1 pixel), most points are shifted closer to the ground truth for both refinement methods, as shown in \cref{fig-test-ref-02b}. The box plots in \cref{fig-test-ref-02c} represent the distribution of ground-truth distances for each bin. In case the upscaled points are already very close to the ground truth (first bin), refinement would not be necessary. Already for the second bin (1 to 2 pixels), we see a clear improvement by the refinement methods. The relative improvement of the refinements increases with the growing ground-truth distances of the upscaled points. For the higher bins, the XR refinement with argmax still achieves a low median of around 2 pixels; however, the IQR is relatively large. For upscaled points with larger ground-truth distances, there may be multiple craquelure locations the refinement method can move the point to, especially for the XR argmax method, which allows more movement flexibility.
Next, we inspect the refinement results of the second modality for the same bins. \cref{fig-test-ref-03a} shows box plots with the ground-truth distances for VIS. The upscaled points are more spread than in \cref{fig-test-ref-02c}. We observe a reduction in ground-truth distances for all refinement methods in the first 7 bins. However, this visualization does not show whether there is a correlation between the refinement in XR and VIS. Thus, we also compare the distance of vector differences to the ground truth of XR and VIS in \cref{fig-test-ref-03b}, which shows a clear improvement for NCC+CN-LoFTR-f for all bins and CN-LoFTR-f for bins 1 to 6.

\begin{figure}
\captionsetup[subfloat]{labelfont=tiny,textfont=tiny}
\centering
\subfloat[][XR GT\label{fig-test-ref-qual-01a}]{%
\includegraphics[width=1.36cm]{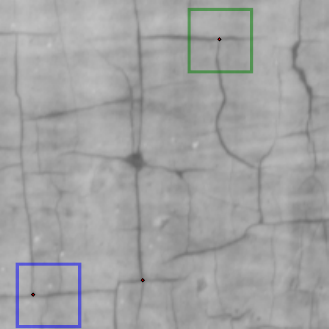}
}
\subfloat[][XR Up\label{fig-test-ref-qual-01b}]{%
\includegraphics[width=1.36cm]{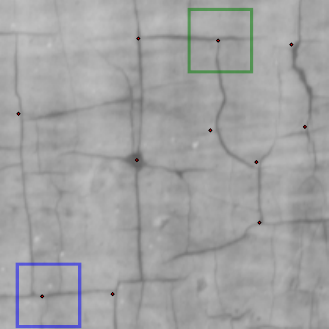}
}
\subfloat[][w/o argmax\label{fig-test-ref-qual-01c}]{%
\includegraphics[width=1.36cm]{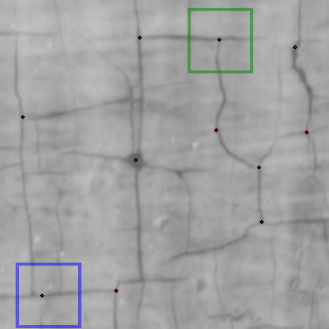}
}
\subfloat[][/w argmax\label{fig-test-ref-qual-01d}]{%
\includegraphics[width=1.36cm]{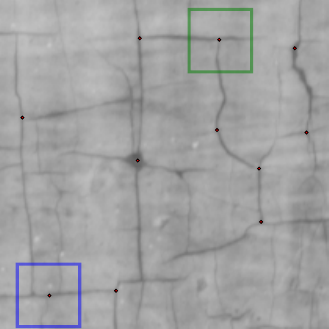}
}
\subfloat[][VIS GT\label{fig-test-ref-qual-01e}]{%
\includegraphics[width=1.36cm]{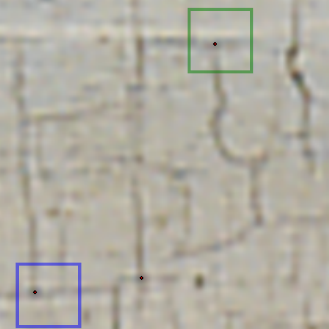}
}
\subfloat[][VIS Up\label{fig-test-ref-qual-01f}]{%
\includegraphics[width=1.36cm]{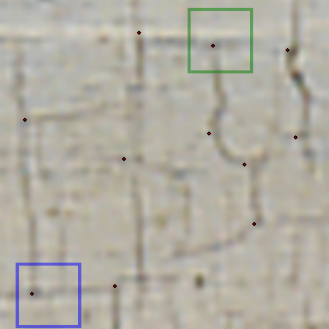}
}

\subfloat[][NCC\\(w.r.t. (c))\label{fig-test-ref-qual-01g}]{%
\includegraphics[width=1.36cm]{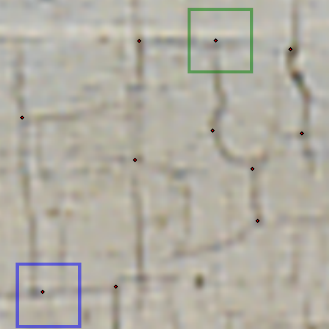}
}
\subfloat[][NCC\\(w.r.t. (d))\label{fig-test-ref-qual-01h}]{%
\includegraphics[width=1.36cm]{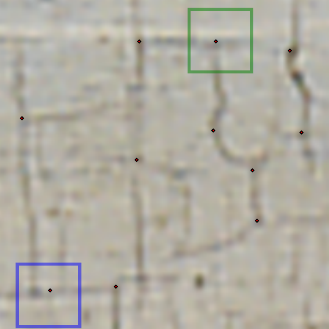}
}
\subfloat[][CN-LoFTR-f\\(w.r.t. (c))\label{fig-test-ref-qual-01i}]{%
\includegraphics[width=1.36cm]{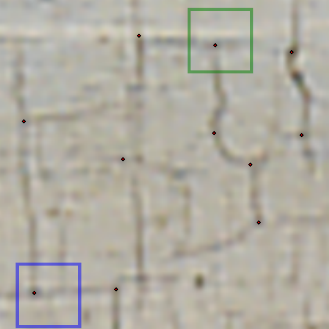}
}
\subfloat[][CN-LoFTR-f\\(w.r.t. (d))\label{fig-test-ref-qual-01j}]{%
\includegraphics[width=1.36cm]{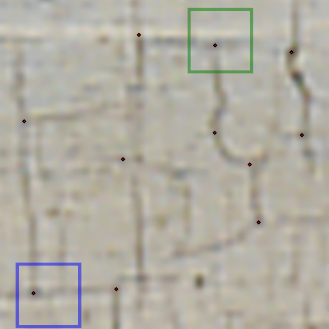}
}
\subfloat[][NCC +\\CN-LoFTR-f \\(w.r.t. (c))\label{fig-test-ref-qual-01k}]{%
\includegraphics[width=1.36cm]{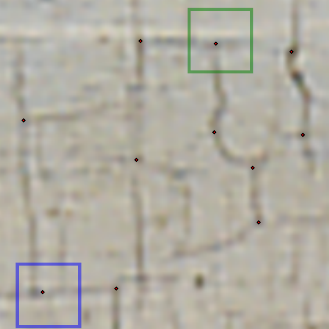}
}
\subfloat[][NCC +\\CN-LoFTR-f \\(w.r.t. (d))\label{fig-test-ref-qual-01l}]{%
\includegraphics[width=1.36cm]{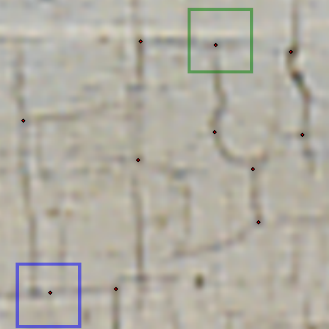}
}

\caption{Qualitative results of keypoint refinement of one example of XR-VIS (Set2 s0) for a tiny image excerpt. The different variants of our refinement method (presented in the ablation study) are visually compared to the upscaling of CN+R+LG keypoints (b,f) and ground-truth control points (a,e). The blue and green rectangles mark two keypoint pairs for a closer inspection of the refinements (XR (c,d) and VIS (g-l)). }
\label{fig-test-ref-qual-01}
\end{figure}

The results for the different combinations of refinement modules are visually compared for a tiny image excerpt of XR-VIS (Set 2 s0) in \cref{fig-test-ref-qual-01}. The upscaled keypoints of CN+R+LG (\cref{fig-test-ref-qual-01b,fig-test-ref-qual-01f}), and the ground-truth control points (\cref{fig-test-ref-qual-01a,fig-test-ref-qual-01e}) are not necessarily at the same crack positions. 
The blue rectangle marks a keypoint pair, where the upscaled point of the first modality (XR) is between two possible crack locations.
Using the XR refinement with argmax (\cref{fig-test-ref-qual-01d}), the point is shifted to the crack junction to the right, while without argmax (\cref{fig-test-ref-qual-01c}), it is only slightly shifted vertically.
For the second modality (VIS), this keypoint initially lies close to the left junction (\cref{fig-test-ref-qual-01f}). Using only NCC (\cref{fig-test-ref-qual-01g,fig-test-ref-qual-01h}), the keypoint is shifted more towards its location in (\cref{fig-test-ref-qual-01c,fig-test-ref-qual-01d}). Using only CN-LoFTR-f (\cref{fig-test-ref-qual-01i,fig-test-ref-qual-01j}), the keypoint cannot move so far and remains at the left junction, thus being a false match. Using both NCC + CN-LoFTR-f (\cref{fig-test-ref-qual-01k,fig-test-ref-qual-01l}), the keypoint is shifted and sub-pixel refined, where the pair in (\cref{fig-test-ref-qual-01d,fig-test-ref-qual-01l}) represents a match centered at a crack junction and in (\cref{fig-test-ref-qual-01c,fig-test-ref-qual-01k}) a match slightly next to the junction. 
For the keypoint pair in the green rectangle, the differences between the detected and ground-truth points are already small, resulting only in minor corrections by the refinement methods.

\begin{figure*}
\scriptsize
\centering
\subfloat[][VIS and XR\label{fig-test-ref-qual-02a}]{
\begin{minipage}[b]{.21\textwidth}
	\includegraphics[width=\textwidth]{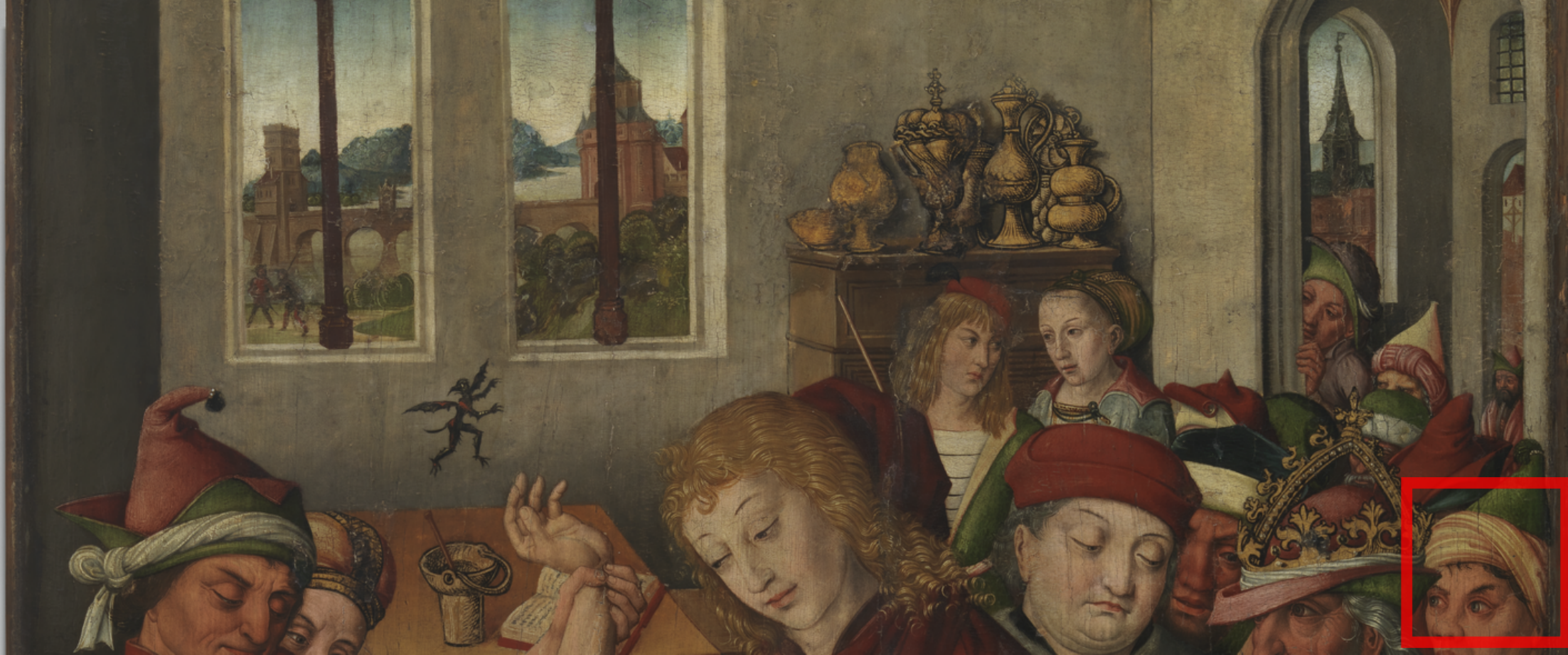}

	\vspace{0.10cm}

	\includegraphics[width=\textwidth]{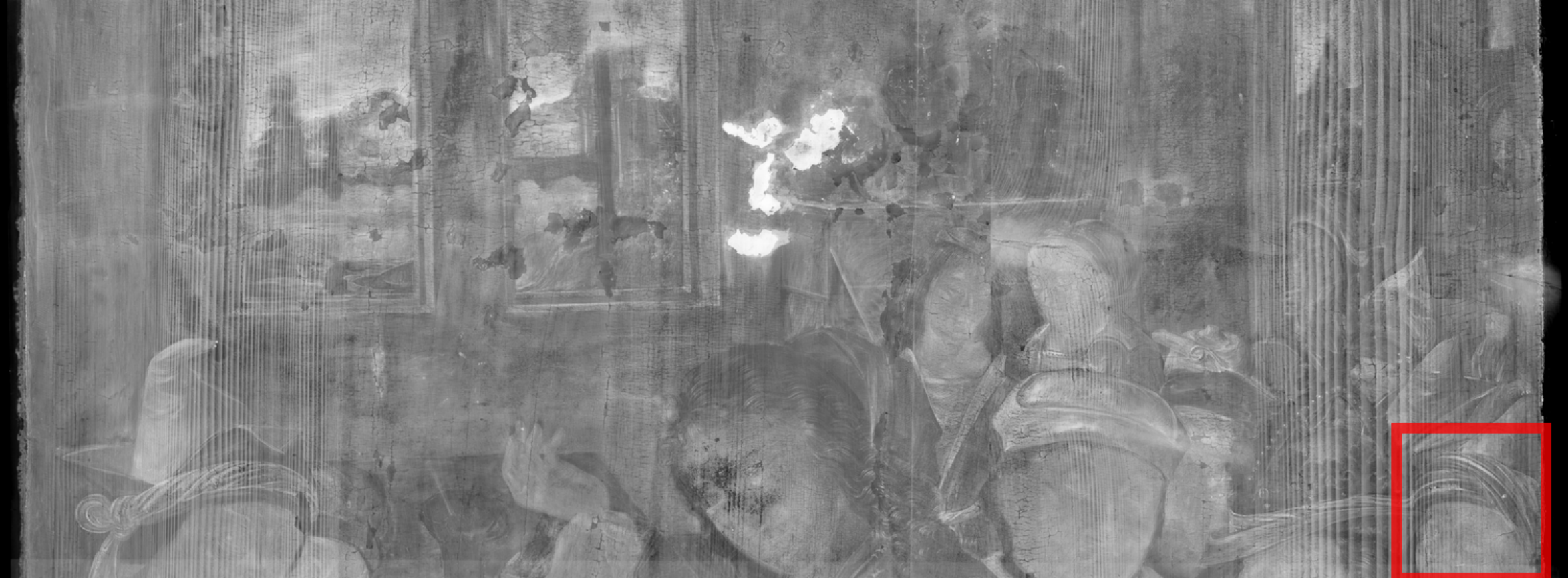}
\end{minipage}
}
\subfloat[][XR\label{fig-test-ref-qual-02b}]{
\includegraphics[width=0.17\textwidth]{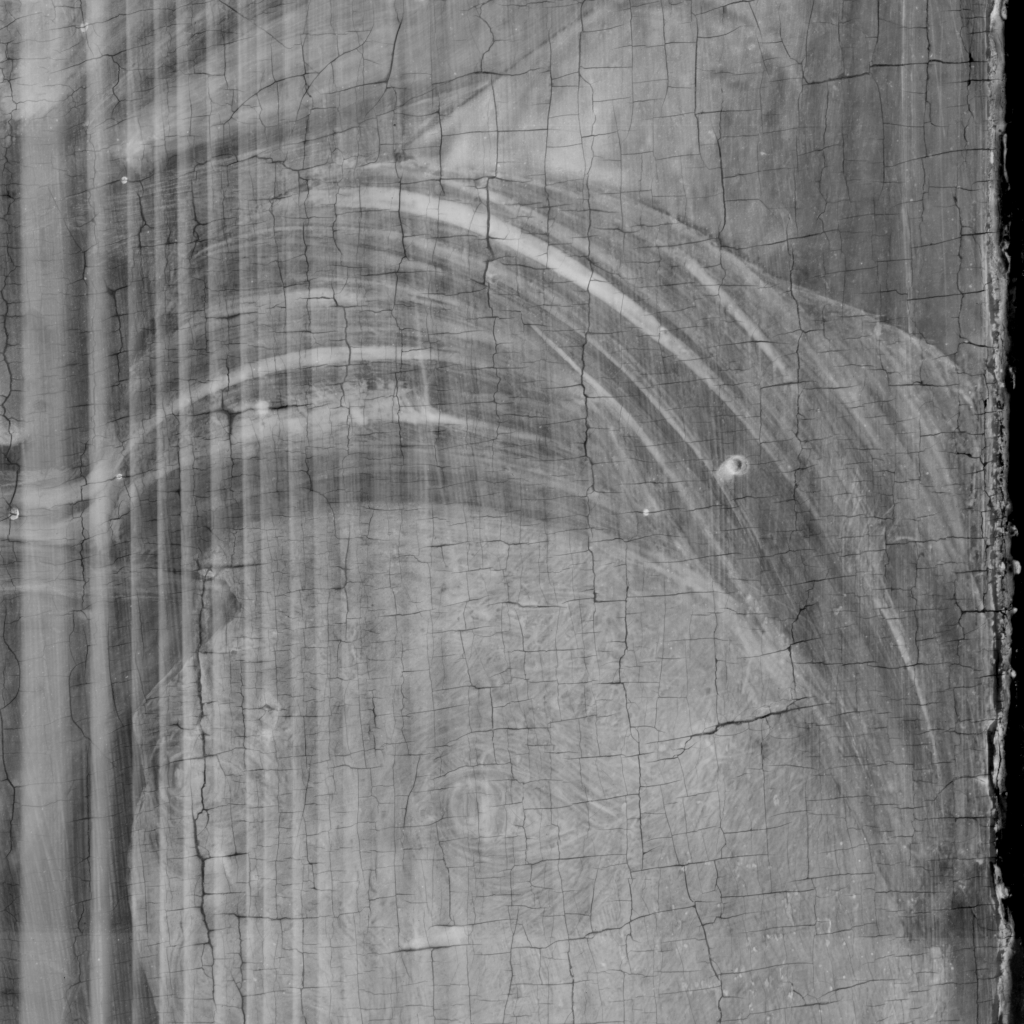}
}
\subfloat[][VIS after TPS\label{fig-test-ref-qual-02c}]{
\includegraphics[width=0.17\textwidth]{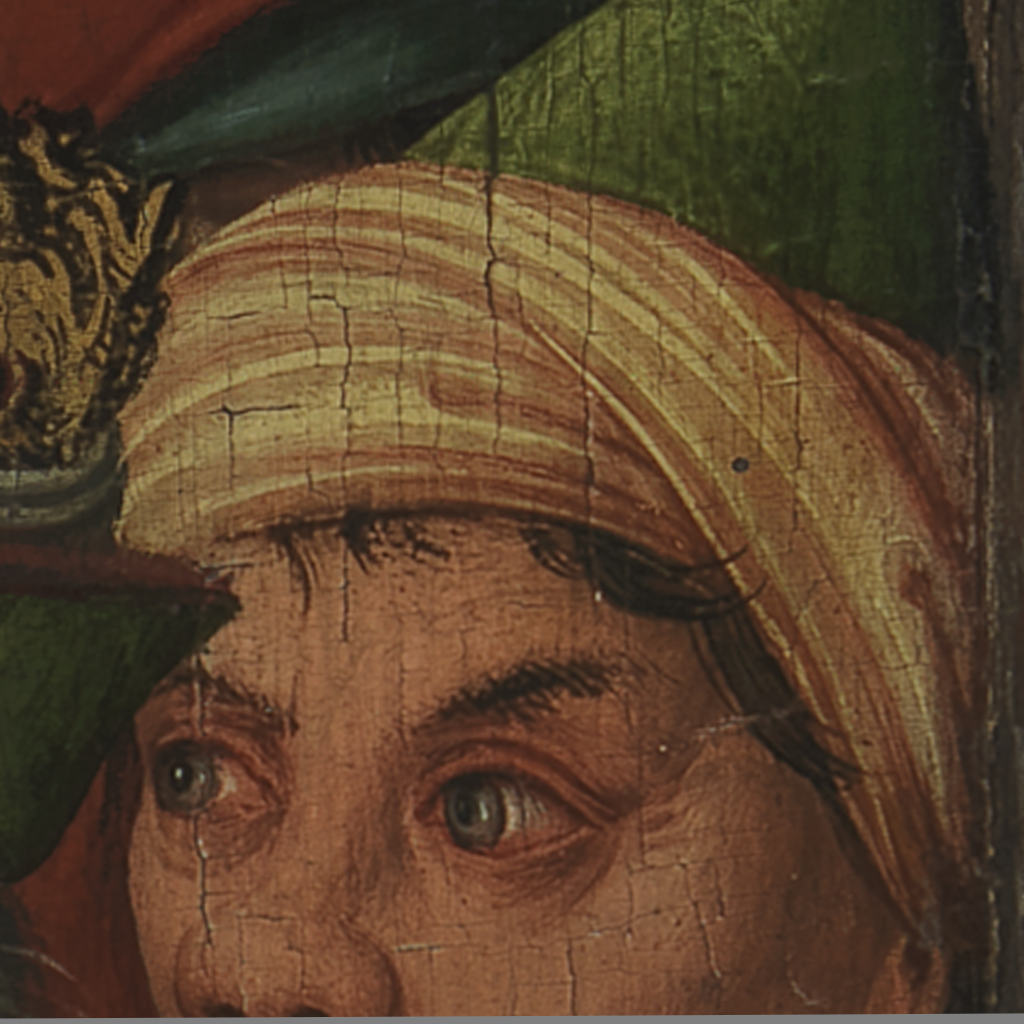}
}
\subfloat[][H-Overlay\label{fig-test-ref-qual-02d}]{
\includegraphics[width=0.17\textwidth]{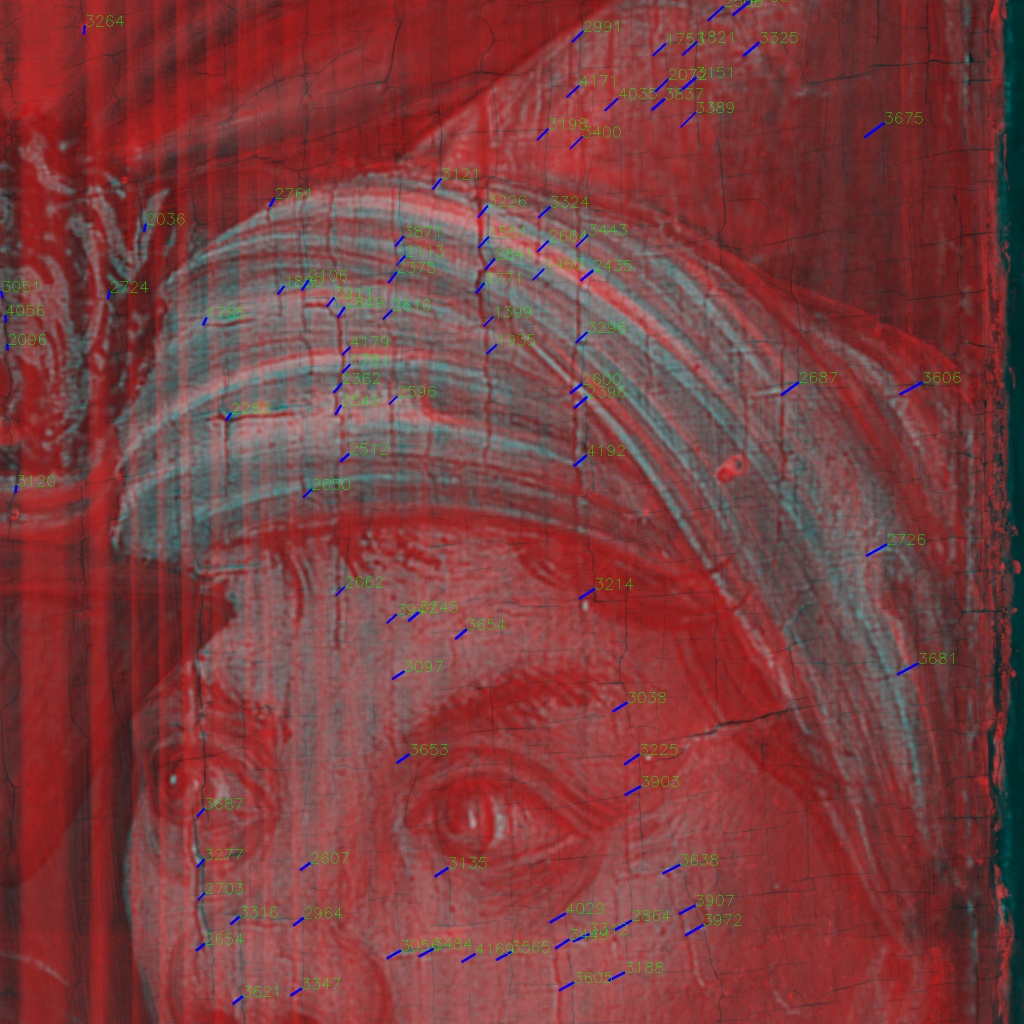}
}
\subfloat[][TPS-Overlay\label{fig-test-ref-qual-02e}]{
\includegraphics[width=0.17\textwidth]{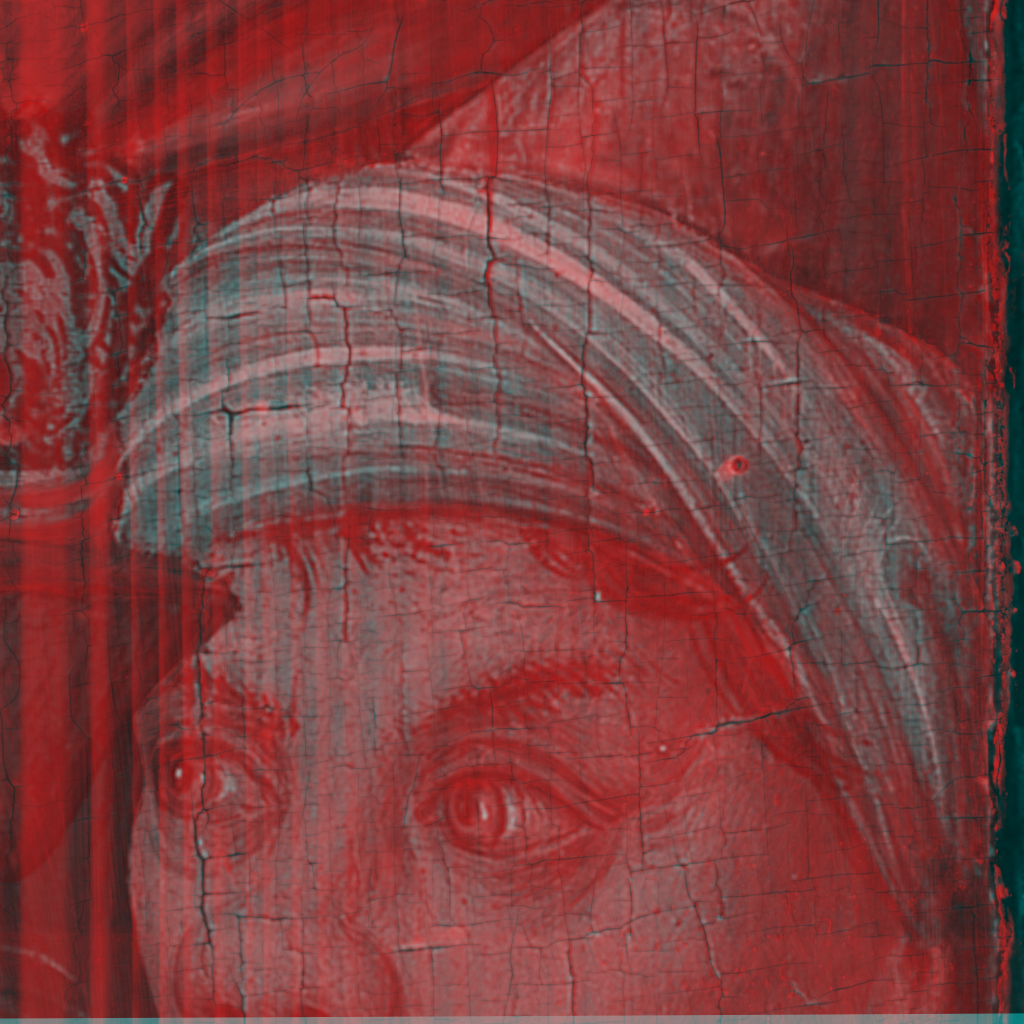}
}
\caption{Qualitative results for one example of XR-VIS Set3 registration using NCC+CN-LoFTR-f. (b-e) are zoom in views of an excerpt to the complete images. (c) represents the registered VIS image using TPS. (d) is the false color overlay of the registered pair using the global homography with superimposed  keypoint displacement vectors. (e) is false color overlay of the non-rigid registered pair. Image sources: Meister des Augustiner-Altars (Hans Traut) und Werkstatt mit Rueland Frueauf d.Ä., Teufelsaustreibung des hl. Veit, VIS and XR, Germanisches Nationalmuseum, Nuremberg, on loan from Evang.-Luth. Kirchengemeinde Nürnberg-St. Lorenz, Gm 146, all rights reserved}
\label{fig-test-ref-qual-02}
\end{figure*}

\begin{table}[t]
\centering
\caption{Quantitative evaluation for XR-VIS (Set1-s0). Best results are highlighted in bold.} 
\label{tab-test-ref-01}
\tiny
\begin{tabular}{l@{\hspace{5pt}}l r@{\hspace{5pt}}r r@{\hspace{5pt}}r r@{\hspace{5pt}}r}
\toprule
 && \multicolumn{6}{c}{XR-VIS (Set1-s0)} \\
Initial & Refinement Methods & \multicolumn{2}{c}{SR-ME [\%] $\uparrow$} & \multicolumn{2}{c}{SR-MAE [\%] $\uparrow$} & ME $\downarrow$ & MAE $\downarrow$ \\
&& $\epsilon=1$ & $\epsilon=2$ & $\epsilon=5$ & $\epsilon=7$ & Mean$\pm$Std & Mean$\pm$Std \\
\midrule
SP+LG-P & Up th2 & 5.0 & 95.0 & 55.0 & \textbf{100.0} & 1.51$\pm$0.29 & 4.58$\pm$1.20 \\
SP+LG-P & KeyPt2SubPx th2 & 5.0 & 95.0 & 60.0 & \textbf{100.0} & 1.51$\pm$0.29 & 4.55$\pm$1.09 \\
\midrule
CN+R+LG & Up th2 & 0.0 & \textbf{100.0} & 65.0 & \textbf{100.0} & 1.50$\pm$0.29 & 4.29$\pm$1.13 \\
CN+R+LG & CN-LoFTR-f th2 & 5.0 & \textbf{100.0} & 70.0 & \textbf{100.0} & \textbf{1.46$\pm$0.29} & \textbf{4.27$\pm$1.09} \\
\bottomrule
\end{tabular}
\end{table}

\begin{table}[t]
\centering
\caption{Quantitative evaluation for XR-VIS (Set2-s0). Best results are highlighted in bold.} 
\label{tab-test-ref-02}
\tiny
\begin{tabular}{l@{\hspace{5pt}}l@{\hspace{5pt}}r@{\hspace{3pt}}r@{\hspace{5pt}}r@{\hspace{3pt}}r@{\hspace{5pt}}r@{\hspace{3pt}}r}
\toprule
 && \multicolumn{6}{c}{XR-VIS (Set2-s0)} \\
Initial & Refinement Methods & \multicolumn{2}{c}{SR-ME [\%] $\uparrow$} & \multicolumn{2}{c}{SR-MAE [\%] $\uparrow$} & ME $\downarrow$ & MAE $\downarrow$ \\
&& $\epsilon=2$ & $\epsilon=3$ & $\epsilon=5$ & $\epsilon=7$ & Mean$\pm$Std & Mean$\pm$Std \\
\midrule
SP+LG-P & Up th3 & 16.7 & 77.8 & 0.0 & 33.3 & 2.75$\pm$1.29 & 13.76$\pm$18.00 \\
SP+LG-P & KeyPt2SubPx th3 & 16.7 & 77.8 & 0.0 & 38.9 & 2.82$\pm$1.66 & 15.17$\pm$21.34 \\
\midrule
CN+R+LG & Up th3 & 16.7 & 88.9 & 5.6 & 44.4 & 2.40$\pm$0.61 & 9.45$\pm$4.43 \\
CN+R+LG & Argmax+NCC+CN-LoFTR-f th3 & 66.7 & \textbf{94.4} & 11.1 & \textbf{61.1} & \textbf{2.02$\pm$0.36} & \textbf{7.12$\pm$2.96} \\
\bottomrule
\end{tabular}
\end{table}

\begin{table}[t]
\centering
\caption{Quantitative evaluation for XR-VIS (Set3-s0). Best results are highlighted in bold.} 
\label{tab-test-ref-03}
\tiny
\begin{tabular}{l@{\hspace{5pt}}l@{\hspace{5pt}}r@{\hspace{3pt}}r@{\hspace{5pt}}r@{\hspace{3pt}}r@{\hspace{5pt}}r@{\hspace{3pt}}r}
\toprule
 && \multicolumn{6}{c}{XR-VIS (Set3-s0)} \\
Initial & Refinement Methods & \multicolumn{2}{c}{SR-ME [\%] $\uparrow$} & \multicolumn{2}{c}{SR-MAE [\%] $\uparrow$} & ME $\downarrow$ & MAE $\downarrow$ \\
&& $\epsilon=3$ & $\epsilon=4$ & $\epsilon=9$ & $\epsilon=11$ & Mean$\pm$Std & Mean$\pm$Std \\
\midrule
SP+LG-P & Up th4 & 15.0 & 60.0 & 10.0 & 30.0 & 3.82$\pm$0.98 & 12.23$\pm$4.03 \\
SP+LG-P & KeyPt2SubPx th4 & 25.0 & 60.0 & 15.0 & 40.0 & 3.75$\pm$0.92 & 12.22$\pm$3.99 \\
\midrule
CN+R+LG & Up th4 & 35.0 & 70.0 & 15.0 & \textbf{60.0} & 3.65$\pm$1.00 & 11.15$\pm$2.65 \\
CN+R+LG & Argmax+NCC+CN-LoFTR-f th4 & 45.0 & \textbf{80.0} & 45.0 & 55.0 & \textbf{3.39$\pm$1.03} & \textbf{10.50$\pm$3.29} \\
\bottomrule
\end{tabular}
\end{table}

\begin{table}[t]
\centering
\caption{Quantitative evaluation for XR-IRR (s0). Best results are highlighted in bold.} 
\label{tab-test-ref-04}
\tiny
\begin{tabular}{l@{\hspace{5pt}}l r@{\hspace{5pt}}r r@{\hspace{5pt}}r r@{\hspace{5pt}}r}
\toprule
 && \multicolumn{6}{c}{XR-IRR (s0)} \\
Initial & Refinement Methods & \multicolumn{2}{c}{SR-ME [\%] $\uparrow$} & \multicolumn{2}{c}{SR-MAE [\%] $\uparrow$} & ME $\downarrow$ & MAE $\downarrow$ \\
&& $\epsilon=2$ & $\epsilon=3$ & $\epsilon=5$ & $\epsilon=7$ & Mean$\pm$Std & Mean$\pm$Std \\
\midrule
SP+LG-P & Up th3 & 27.8 & 94.4 & 11.1 & 44.4 & 2.22$\pm$0.34 & 8.45$\pm$3.13 \\
SP+LG-P & KeyPt2SubPx th3 & 16.7 & 94.4 & 0.0 & 27.8 & 2.27$\pm$0.30 & 8.72$\pm$3.03 \\
\midrule
CN+R+LG & Up th3 & 33.3 & \textbf{100.0} & 0.0 & \textbf{55.6} & 2.11$\pm$0.23 & 7.69$\pm$2.99 \\
CN+R+LG & CN-LoFTR-f th3 & 44.4 & \textbf{100.0} & 16.7 & \textbf{55.6} & \textbf{2.06$\pm$0.30} & \textbf{7.47$\pm$2.84} \\
\bottomrule
\end{tabular}
\end{table}

Based on the ablation study findings, we select one refinement method for each dataset: for XR-VIS Set1 and XR-IRR, we choose XR without argmax and CN-LoFTR-f; and for XR-VIS Set2-3, we choose XR with argmax and NCC + CN-LoFTR-f. The results after outlier filtering and TPS computation are shown in~\cref{tab-test-ref-01,tab-test-ref-02,tab-test-ref-03,tab-test-ref-04}. For XR-VIS Set1 ($th_{out}=2$), the registration results yield small improvements for our refinement method compared to upscaling. Especially, SR-ME is superior to SP-KeyPt2SubPx refinement and upscaling. The only moderate improvements are unsurprising, given that we only have a small scaling factor, such that the initial results are expected to be very good.
For XR-IRR ($th_{out}=3$), our refinement method slightly reduces ME and MAE compared to SP-KeyPt2SubPx. 
Here, the scaling factor is higher than for XR-VIS Set1, but still, the initial points within the 3-pixel outlier filtering are already quite good.
For XR-VIS Set2 ($th_{out}=3$), our refinement method using the complete building blocks reduces the mean ME and mean MAE distinctly with a gain of nearly 5\% in SR-ME at 3 pixels and of nearly 17\% in SR-MAE at 7 pixels. SP-KeyPt2SubPx only shows improvements in SR-MAE at 7 pixels, but on a lower basis. The larger scaling factor magnifies errors of the initial keypoints during upscaling, such that the refinement methods have a larger effect.
For XR-VIS Set3 ($th_{out}=4$), our refinement method improves SR-ME, mean ME, and mean MAE. SP-KeyPt2SubPx shows improvements for SR-MAE and mean ME, but does not yield the performance of our method. Due to the large scaling factor of up to 6.5, this dataset is the most challenging one for refinement. The bicubically upsampled VIS images are blurry, while the XR images are sharp. Hence, the exact placement of the same features in the crack structure is difficult, as the borders of the cracks in the VIS image are smeared out. 

\cref{fig-test-ref-qual-02} shows an example for XR-VIS Set3 registration, where the scaling factor between original VIS and XR resolution is about 4.5. The visual image shows large distortions in the border areas. In the false-color overlay after global homography estimation (\cref{fig-test-ref-qual-02d}), this manifests in large displacement vectors at the distorted areas. The false-color overlay after TPS computation (\cref{fig-test-ref-qual-02e}) shows clearly that the distortion in these areas has been corrected by our method.

\subsection{Limitations}
The proposed method was designed for the multi-modal registration using crack characteristic features, such as branching points or sharp bends. For that reason, the keypoint detector will be most effective for images with crack-like structures and is also transferable to, \eg, vessels~\cite{SindelA2022BVM,SindelA2022MICCAI}, but is not intended to work well for computer vision datasets with only plain objects. 
Our patch-based non-rigid approach requires images with similar scaling factors and image areas, and slight rotation. If these conditions are not met, a global prealignment step must be added, which could be performed at low resolution, to define the regions for patch-based registration. 

\section{Conclusion}
We propose a coarse-to-fine non-rigid multi-modal registration method for panel paintings that addresses registration of large images, varying image resolutions, and non-rigid distortions by applying an efficient patch-based method using keypoints, a sparse deep matcher, and TPS, thereby reducing complexity compared to dense displacement-field methods. Keypoints are filtered within local areas using homographies to remove outliers. With crack-based keypoints as control points for the TPS, we directly control the deformation and make it traceable. The multi-level keypoint refinement modules enable upscaling by up to a factor of 4 and have the potential to be applied to other domains with large high-resolution images. 
With an extensive test set of panel paintings, we numerically showed the effectiveness of our one-stage and coarse-to-fine registration methods.

\bibliographystyle{IEEEtran}
\bibliography{refs}

\begin{IEEEbiographynophoto}{Aline Sindel} received the B.Sc. degree and the M.Sc. degree in Medical Engineering from the Friedrich-Alexander-Universität Erlangen-Nürnberg, Erlangen, Germany, in 2015 and 2018, respectively. She is currently pursuing her PhD degree in computer science at the Pattern Recognition Lab, Friedrich-Alexander-Universität Erlangen-Nürnberg, where she specializes in multi-modal image registration. Her research interest includes machine learning, computer vision, and image processing techniques for applications in art and medical imaging.
\end{IEEEbiographynophoto}
\begin{IEEEbiographynophoto}{Andreas Maier}
	(Senior Member, IEEE) graduated in computer science and the Ph.D. degree from Friedrich-Alexander-Universität Erlangen-Nürnberg (FAU), Erlangen, Germany, in 2005 and 2009, respectively. 
	In this period, his major research subject was medical signal processing in speech data. 
	From 2009 to 2010, he started working on flat-panel C-arm CT as a Post-Doctoral Fellow at the Radiological Sciences Laboratory, Stanford University, Stanford, CA, USA. 
	From 2011 to 2012, he was with Siemens Healthcare, Erlangen, Germany, as the Innovation Project Manager and was responsible for reconstruction topics for angiography and X-ray. 
	In 2012, he returned to FAU as the Head of the Medical Reconstruction Group, Pattern Recognition Lab, where he became a Professor and the Head in 2015. His research interests include medical imaging, image and audio processing, digital humanities, and interpretable machine learning and the use of known operators. He has been a member of the Steering Committee of the European Time Machine Consortium since 2016. In 2018, he received the ERC Synergy Grant ``4D nanoscope".
\end{IEEEbiographynophoto}
\begin{IEEEbiographynophoto}{Vincent Christlein}
	received the degree in computer science and the Ph.D. (Dr.-Ing.) degree from Friedrich-Alexander-Universität Erlangen-Nürnberg (FAU), Erlangen, Germany, in 2012 and 2018, respectively. During his studies, he worked on automatic handwriting analysis with a focus on writer identification and writer retrieval. Since 2018, he has been a Research Associate with the Pattern Recognition Lab, FAU, where he was promoted to an Academic Councilor in 2020 and heads the Computer Vision Group, which covers a wide variance of topics, e.g., computational humanities topics, such as document and art analysis, but also environmental projects, such as glacier segmentation or solar cell crack recognition.
\end{IEEEbiographynophoto}
\vfill

\end{document}